\documentclass[12pt]{article}
\usepackage{amsmath}
\usepackage{graphicx}
\usepackage{enumerate}
\usepackage{natbib}
\usepackage{url}

\newcommand{\blind}{1}
\usepackage{amsmath}
\usepackage{bbm}
\usepackage{graphicx,psfrag,epsf}
\usepackage{enumerate}
\usepackage{natbib}
\usepackage{amsfonts}
\usepackage{amsthm}
\usepackage{amssymb}
\usepackage{bm}
\usepackage{caption}
\usepackage{multirow}
\usepackage{booktabs}
\usepackage{url} 
\usepackage{color}
\usepackage[colorlinks,citecolor=blue,urlcolor=blue]{hyperref}
\usepackage{graphicx}
\usepackage{caption}
\usepackage{subcaption}
\usepackage{comment}
\usepackage{float}     
\usepackage{placeins} 
\usepackage{amsfonts}
\usepackage{makecell}

\usepackage{xr}
\makeatletter
\newcommand*{\addFileDependency}[1]{
	\typeout{(#1)}
	\@addtofilelist{#1}
	\IfFileExists{#1}{}{\typeout{No file #1.}}
}
\makeatother

\newcommand*{\myexternaldocument}[1]{%
	\externaldocument{#1}%
	\addFileDependency{#1.tex}%
	\addFileDependency{#1.aux}%
}
\myexternaldocument{supp}

	\newtheorem{theorem}{Theorem}

\begin{document}
\def\spacingset#1{\renewcommand{\baselinestretch}%
{#1}\small\normalsize} \spacingset{1}
\def\var{{\rm Var}\,}
	\def\blue#1{\textcolor{blue}{#1}}
	\def\red#1{\textcolor{red}{#1}}

\if1\blind
{
  \title{\bf Online federated learning framework for classification}
		\author{Wenxing Guo\thanks{
				Co-first author }\hspace{.2cm}\\
			School of Mathematics, Statistics and Actuarial Science,\\ University of Essex,
Colchester, CO4 3SQ, United Kingdom \vspace{0.3cm}\\
			Jinhan Xie$^{*}$
            \hspace{.2cm}\\
			Yunnan Key Laboratory of Statistical Modeling and Data Analysis, \\Yunnan University, Kunming, 650091, China \vspace{0.3cm}\\
            Jianya Lu$^{*}$
             \hspace{.2cm}\\
             School of Mathematics, Statistics and Actuarial Science,\\ University of Essex,
Colchester, CO4 3SQ, United Kingdom \vspace{0.3cm}\\
             Bei Jiang
             \hspace{.2cm}\\
Department of Mathematical and Statistical Sciences,\\
University of Alberta,
Edmonton, AB, T6G 2G1, Canada 
            \vspace{0.3cm}\\
             Hongsheng Dai \hspace{.2cm}\\
School of Mathematics, Statistics and Physics,\\ Newcastle University, Newcastle upon Tyne, NE1 7RU, United Kingdom 
             \vspace{0.3cm}\\
             Linglong Kong
             \hspace{.2cm}\\
Department of Mathematical and Statistical Sciences,\\
University of Alberta,
Edmonton, AB, T6G 2G1, Canada 
			}
		\maketitle
	} \fi

\begin{abstract}
 In this paper, we develop a novel online federated learning framework for classification, designed to handle streaming data from multiple clients while ensuring data privacy and computational efficiency. Our method leverages the generalized distance-weighted discriminant technique, making it robust to both homogeneous and heterogeneous data distributions across clients. 
In particular, we develop a new optimization algorithm based on the Majorization-Minimization principle, integrated with a renewable estimation procedure, enabling efficient model updates without full retraining. We provide a theoretical guarantee for the convergence of our estimator, proving its consistency and asymptotic normality under standard regularity conditions. In addition, we establish that our method achieves Bayesian risk consistency, ensuring its reliability for classification tasks in federated environments.
We further incorporate differential privacy mechanisms to enhance data security, protecting client information while maintaining model performance. Extensive numerical experiments on both simulated and real-world datasets demonstrate that our approach delivers high classification accuracy, significant computational efficiency gains, and substantial savings in data storage requirements compared to existing methods. 
\end{abstract}

\noindent%
{\it Keywords:} Streaming data; Distance-weighted discriminant; Online learning;  Privacy preservation; Computational efficiency.

\spacingset{1.9}

\section{Introduction}
The exponential growth in data generated by end-user devices (also called clients), including mobile phones, wearable technologies, and autonomous vehicles, has garnered significant attention. Advances in the storage and computational capacities of these devices, coupled with growing concerns about data privacy, have driven significant interest in Federated Learning (FL) \citep{mcmahan2017communication, kairouz2021advances, wu2022communication}. In a typical FL system, a large number of clients collaborate to train a shared global model while ensuring that the data are stored locally on each device. Since its inception, FL has proven to provide notable advantages in both model accuracy and computational efficiency, due to the participation of numerous clients and vast data sources \citep{basu2019qsparse, dai2019hyper, sattler2019robust, fallah2020personalized, wei2020federated, zhu2020federated, li2023targeting, liu2024robust, zhou2024federated}. However, as discussed in \cite{bonawitz2019towards, li2022statistical}, the continuous arrival of data from client devices (e.g., mobile phones) in an online setting presents new challenges that may lead to failure on traditional FL systems, which are typically designed for static datasets stored on each device.

Most of the aforementioned FL methods based on batch learning face two significant limitations when applied to streaming data, particularly in terms of scalability and memory efficiency. One is that retraining the model from scratch whenever new data is introduced leads to delayed model updates, increased computational overhead, and the risk of using outdated models. Another significant drawback is its inability to effectively manage concept drift, where the statistical properties of the data evolve over time. Until now, online learning methods have been widely investigated in both statistics and machine learning, including the aggregated
estimating equation \citep{lin2011aggregated, schifano2016online}, the stochastic gradient descent (SGD) algorithm and its variants \citep{robbins1951stochastic, chen2020statistical, xie2023scalable, han2024online}, as well as renewable estimation and incremental inference \citep{luo2020renewable, han2022inference, luo2023statistical}. 

In the online FL literature, several studies have focused on designing variants of the SGD algorithms. For example, \cite{agarwal2018cpsgd} designed a cpSGD algorithm, a communication-efficient method for distributed mean estimation with privacy guarantee. \citep{yuan2020federated} proposed federated accelerated SGD (FEDAC), a principled acceleration of federated averaging for distributed optimization.
\cite{li2022soteriafl} introduced a unified framework that improves communication efficiency in private FL through communication compression. \cite{koloskova2022sharper} considered an asynchronous SGD algorithm for distributed training. \cite{li2022statistical} studied how to perform statistical inference via Local SGD in FL.
\cite{glasgow2022sharp} provided sharp lower bounds for homogeneous and heterogeneous federated averaging (Local SGD). \cite{guo2022hybrid} investigated a optimization algorithm called HL-SGD for FL with heterogeneous communications. While these existing studies primarily focus on designing algorithms to enhance communication efficiency, the development of methods that address the statistical properties  in FL settings with streaming data is largely lacking. 
In addition to statistical considerations, privacy concerns are another crucial aspect of FL, as decentralized learning involves sensitive user data distributed across multiple devices.

To enhance data privacy protection, we have incorporated differential privacy (DP) techniques in this study.
Differential privacy was first introduced by \cite{dwork2006calibrating}, and this pioneering work has since been widely applied in numerous privacy-related studies. 
For example, \cite{kasiviswanathan2011can} developed algorithms and bounds for private learning problems and showed the equivalence of local private learning and statistical query learning. \cite{abadi2016deep} developed DP-SGD, an algorithm that integrates DP into deep learning via gradient perturbation. 
Differentially private algorithms for convex empirical risk minimization were studied by \cite{chaudhuri2011differentially} and \cite{kifer2012private}, focusing on methods like output perturbation and objective perturbation to ensure privacy.
Recent studies have also examined DP in federated learning (FL). \cite{koloskova2020unified} explored decentralized stochastic optimization and provided a unified convergence analysis for various decentralized SGD methods. 
\cite{cai2023private} developed differentially private methods for high-dimensional linear regression, including DP-BIC for model selection and a debiased LASSO algorithm.
In addition, Recent works have connected DP with federated learning to address the privacy challenges, such as
\cite{dubey2020differentially} introduced a differentially-private federated learning algorithm for contextual linear bandits.
\cite{liu2022privacy} proposed MR-MTL as an effective personalization baseline that enhances collaboration under privacy constraints while analyzing its theoretical and empirical impact on data heterogeneity.
\cite{allouah2023privacy} analyzed the fundamental trade-off between privacy, robustness and utility in distributed ML, and proposed an algorithm that ensures both differential privacy and robust aggregation against adversarial behaviors.
\cite{zhang2024differentially} proposed a federated estimation algorithm for linear regression and developed statistical inference methods with differential privacy,  
\cite{li2024federated} addressed data heterogeneity and privacy in federated transfer learning by formulating federated differential privacy.

\begin{figure}[h]  
  \centering  
  \includegraphics[width=0.8\textwidth]{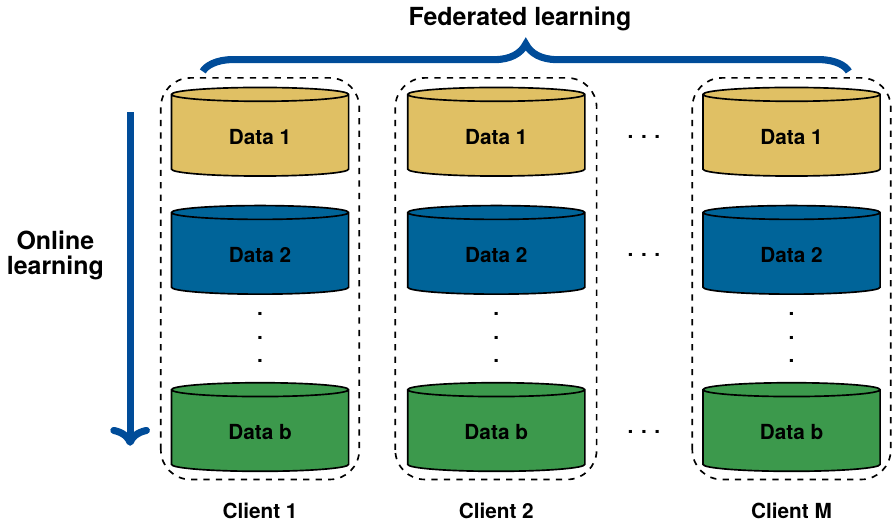}  
  \caption{A diagrammatic representation of online and federated learning frameworks}  
  \label{intro:1-1}  
\end{figure}

The article focuses on the challenge of online learning with privacy preservation in classification problems. Our study is motivated by the prevalence of real-world datasets that consist of streaming data from multiple clients, particularly within federated learning systems that handle massive amounts of non-IID data. As shown in Figure \ref{intro:1-1},  our goal in this work is to develop a method to handle streaming data from multiple non-communicating clients within the framework of online learning and federated learning. 
The key contributions of our work and their significance are outlined as follows:
\begin{itemize}
    \item We address the classification problem for streaming data within the federated learning paradigm by proposing a novel online learning framework that integrates privacy-preserving techniques. In addition, we introduce an offline learning approach that ensures data security while efficiently handling both homogeneous and heterogeneous data distributions. Our methods are specifically designed to mitigate the challenges posed by non-IID data across multiple clients while preserving performance stability and robustness.
    \item We develop a novel optimization algorithm based on the Majorization-Minimization (MM) principle, enhanced with a renewable estimation procedure introduced by \cite{luo2020renewable}. To establish the reliability of our approach, we rigorously derive a comprehensive convergence theory and demonstrate that the proposed estimator, $\tilde{\boldsymbol{\theta}}_b$, is both consistent and asymptotically normal under standard assumptions. Furthermore, we provide a formal proof verifying that our classification method achieves Bayesian risk consistency, ensuring its effectiveness in practical applications.
    \item We establish the differential privacy guarantees of our proposed method. We show that our online federated learning algorithm satisfies $\epsilon$-differential privacy when noise is drawn from the Laplace distribution and $(\epsilon, \delta)$-differential privacy when noise follows a Gaussian distribution. This ensures that our method provides strong privacy protection while maintaining high classification accuracy, making it a reliable solution for privacy-sensitive applications.
    \item Extensive numerical experiments on both simulated and real-world datasets validate the efficacy of our method. Our approach not only exhibits superior computational efficiency but also significantly reduces data storage overhead compared to the generalized distance-weighted discriminant method. Furthermore, when applied to real-world data, our model consistently demonstrates enhanced stability, improved predictive performance, and remarkable computational advantages, highlighting its practical significance and applicability.
\end{itemize}

\section{Preliminary and problem setup}

In this section, we briefly
review some background knowledge about distance-weighted discriminant and outlines the problem setup.
\subsection{Distance-weighted discriminant}
The distance-weighted discriminant (DWD) method, introduced by Marron et al. \citep{marron2007distance}, aims to identify a separate hyperplane that minimizes the sum of the inverse margins of all data points. 
The DWD method is a classification technique that overcomes several limitations of traditional classifiers, particularly in imbalanced, high-dimensional, and noisy data settings.
DWD optimizes margin size by considering the reciprocal of distances rather than just focusing on support vectors, leading to more robust and balanced decision boundaries. By incorporating distance-weighting, DWD reduces the influence of noisy or misclassified data points, leading to higher classification accuracy and stability.
Given a training dataset consisting of $n$ pairs of observations, $\left\{\bm x_i, y_i\right\}_{i=1}^n$, $\bm x_i \in \mathbb{R}^p$ and $y_i \in\{-1,1\}$, the linear DWD seeks a hyperplane defined by $\left\{\bm x: \bm x^{\top} \hat{\boldsymbol{\beta}}+\hat{\beta}_0=0\right\}$. The optimization problem is expressed as:
\begin{eqnarray}
(\hat{\boldsymbol{\beta}}, \hat{\beta}_0)={\rm{\mathop {\min }\limits_{\bm\beta, \beta_0}}}
\bigg(\sum_{i=1}^n\frac{1}{r_i}+ C\sum_{i=1}^n\eta_i\bigg),
\label{1-1}
\end{eqnarray}
subject to the constraints $r_i=y_i(\bm {x_i}^T\bm{\beta}+ \beta_0)+ \eta_i$, 
$r_i\ge 0$, $\eta_i\ge 0$ and $\bm{\beta}^T\bm{\beta}=1$, where $C$ is a tuning parameter that regulates the slack variables $\eta_i$. 
\\
\cite{wang2018another} proposed an algorithm for solving generalized DWD, and demonstrated that the generalized DWD approach achieves good classification accuracy and reduced computation time compared to SVM.

The generalized DWD, replacing the reciprocal in the standard DWD optimization problem (\ref{1-1})
with the $q$th power $(q>0)$ of the inverse distances, is defined as:
\begin{eqnarray}
{\rm{\mathop {\min }\limits_{\bm\beta, \beta_0}}}
\bigg(\sum_{i=1}^n\frac{1}{r_i^q}+ C\sum_{i=1}^n\eta_i\bigg).
\label{1-2}
\end{eqnarray}
Let 
\begin{eqnarray}
L(\bm\beta, \beta_0)=
\sum_{i=1}^nV_q\Big(y_i(\bm {x_i}^T\bm{\beta}+ \beta_0)\Big)+ \frac{n\lambda}{2}\bm\beta^T\bm\beta,
\label{1-4}
\end{eqnarray}
where
\begin{equation}
V_q(u)=\left\{
\begin{array}{rcl}
1-u, ~~~~~&& {{\rm if}}~ u\le \frac{q}{q+1},\\
\frac{1}{u^q}\frac{q^q}{(q+1)^{q+1}}, && {{\rm if}}~ u> \frac{q}{q+1}.
\end{array} \right.
\label{1-5}
\end{equation} 
The generalized DWD classifier (\ref{1-2}) can be expressed as $\operatorname{sign}(\bm x_i^T\hat{\bm\beta}+\hat\beta_0)$. \cite{wang2018another} showed that $(\hat{\bm\beta}, \hat\beta_0)$ can be obtained by minimizing the objective function
\begin{eqnarray}
\frac{1}{n}\sum_{i=1}^nV_q\Big(y_i(\bm {x_i}^T\bm{\beta}+ \beta_0)\Big)+ \frac{\lambda}{2}\bm\beta^T\bm\beta.
\label{1-6}
\end{eqnarray}
In addition, let $\bm\theta=(\beta_0, \bm\beta^T)^T$, we have
\begin{eqnarray}
L(\bm\theta)=
\sum_{i=1}^nV_q(y_i\bar{\bm{x}}_i^T\bm{\theta})+ \frac{n\lambda}{2}\bm\theta^T\bm W\bm\theta,
\label{1-4-1}
\end{eqnarray}
where $\bar{\bm{x}}_i^T=(1,{\bm{x}}_i^T)$ and  
$\bm W=\Bigg(\begin{array}{cc}0 & \bm{0}^T \\ \bm{0} & \bm I_p\end{array}\Bigg)$, $\bm{0}$ denotes a column vector consisting of $p$ zeros.

\subsection{Notation and Problem setup}
Throughout the paper, we use $\|\bm x\|$ to denote the $L_2$-norm of a vector $\bm x$.
For sequences $a_n$ and $b_n$, the notation $a_n=o\left(b_n\right)$ implies $a_n / b_n \rightarrow 0$ as $n$ increases, while $a_n=O\left(b_n\right)$ indicates that $a_n\leq C b_n$ for some positive constant $C$. Similarly, $a_n=O_p\left(b_n\right)$ denotes that $a_n / b_n$ is bounded in probability.
We assume that there are $b$ batches of streaming data $D_1, D_2, \ldots, D_b$, where $D_j$ arrives after $D_{j-1}$. In addition, there exists a central server and $M$ local clients.
We denote the data on these clients for the $j$th batch by $D_j^{\scriptscriptstyle(1)}, D_j^{\scriptscriptstyle(2)}, \ldots, D_j^{\scriptscriptstyle(M)}$, respectively.
For the $j$th batch, $j=1,\ldots, b$ and $m$th client, $m=1,2, \ldots, M$, there are $n_j^{\scriptscriptstyle(m)}$ data points. Let $L^m_{D_j}$ be the function $L$ for the $m$th client of the $j$th batch and $N_j$ be the total number of data points of the $j$th batch, i.e., $N_j=\sum_{m=1}^Mn_j^{(m)}$.
The objective of this paper is to develop a robust classification method capable of handling streaming data from multiple non-communicating clients within the framework of both online learning and federated learning, ensuring both computational efficiency and data privacy.

\section{Federated learning}
\begin{figure}[htbp]
    \centering
    \begin{subfigure}[b]{0.47\textwidth}
        \centering
        \includegraphics[width=\textwidth]{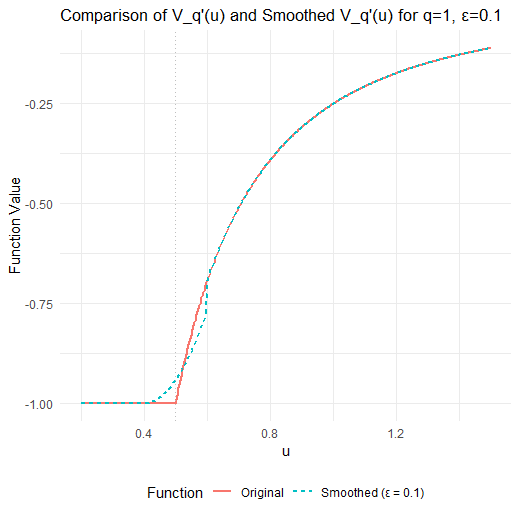}
        \caption{
}
        \label{fig:sub1}
    \end{subfigure}
    \hfill
    \begin{subfigure}[b]{0.47\textwidth}
        \centering
        \includegraphics[width=\textwidth]{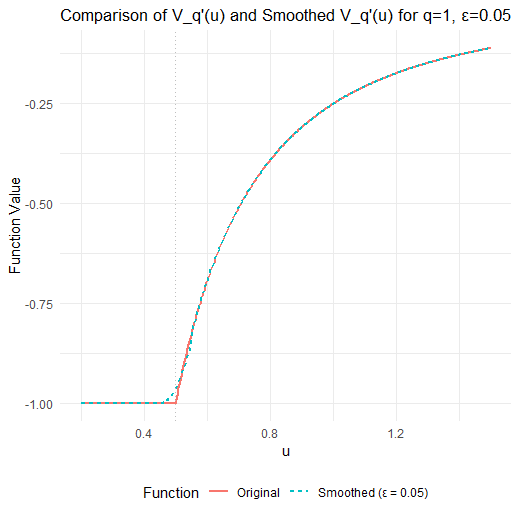}
        \caption{}
        \label{fig:sub2}
    \end{subfigure}
    \vfill
     \caption{Comparison of $V_q'(u)$ and Smoothed $V_q'(u)$}
    \label{figS}
\end{figure}

FL is an efficient and privacy-preserving approach that enables collaborative model training across distributed clients without requiring real-time communication. Its ability to handle heterogeneous data distributions while reducing communication costs makes it an ideal choice for applications in sensitive and resource-constrained environments.
\\
For the data from the dataset $D$. To simplify the notation, we remove the subscript of $L_D^m$ without loss of generality. we consider an approximation $\tilde L^m(\bm\theta)$ of $L^m(\bm\theta)$ by its the second-order expansion of $\bm\theta^0$, 
\begin{eqnarray}
\tilde L^m(\bm\theta)= L^m(\bm\theta^0)+ \sum_{m=1}^M(\bm\theta-\bm\theta^0)^T\nabla L^m(\bm\theta^0)+ \frac{1}{2}\sum_{m=1}^M(\bm\theta-\bm\theta^0)^TH^m(\bm\theta^0)(\bm\theta-\bm\theta^0),
\label{2-1}
\end{eqnarray}
where 
\begin{eqnarray}\nonumber
\nabla L^m(\bm\theta^0)=\sum_{i=1}^{n^{\scriptscriptstyle(m)}}y_iV_q'(y_i\bar{\bm{x}}_i^T\bm{\theta^0})\bar{\bm{x}}_i+  n^{\scriptscriptstyle(m)}\lambda \bm W\bm\theta^0,
\label{2-1-1}
\end{eqnarray}
\begin{eqnarray}\nonumber
H^m(\bm\theta^0)=\sum_{i=1}^{n^{\scriptscriptstyle(m)}}\tilde V_q''(y_i\bar{\bm{x}}_i^T\bm{\theta^0})\bar{\bm{x}}_i\bar{\bm{x}}_i^T+ n^{\scriptscriptstyle(m)}\lambda \bm W
\label{2-1-2}
\end{eqnarray}
and
\begin{equation}\nonumber
V_q^{\prime}(u)= \begin{cases}-1, & \text { if } u \leqslant \frac{q}{q+1}, \\ -\frac{1}{u^{q+1}}\left(\frac{q}{q+1}\right)^{q+1}, & \text { if } u>\frac{q}{q+1},\end{cases}
\label{1-51}
\end{equation}
Since the function $V_q^{\prime}(u)$ is continuous but not differentiable at the point
$u_0=\frac{q}{q+1}$,
the following function is used to approximates $V_q^{\prime}(u)$ in this work. 
$$
\tilde{V}_q^{\prime}(u)= \begin{cases}-1, & \text { if } u \leq u_0-\varepsilon \\ a\left(u-u_0\right)^2+b\left(u-u_0\right)+c, & \text { if } u_0-\varepsilon<u<u_0+\varepsilon \\ -\frac{1}{u^{q+1}}\left(\frac{q}{q+1}\right)^{q+1}, & \text { if } u \geq u_0+\varepsilon\end{cases}
$$
where $a=\frac{(q+1)u_0^{q+1}}{4\varepsilon(u_0+\varepsilon)^{q+2}}$, $b=\frac{(q+1)u_0^{q+1}}{2(u_0+\varepsilon)^{q+2}}$ and $c=-1+\frac{\varepsilon(q+1)u_0^{q+1}}{4(u_0+\varepsilon)^{q+2}}.$
\\
As shown in Figure \ref{figS}, this smoothed function $\tilde{V}_q^{\prime}(u)$ is differentiable and closely approximates $V_q^{\prime}(u)$ for small $\varepsilon$.

Thus, we obtain the approximate second derivative of function $V_q(u)$.
\begin{eqnarray}
\tilde{V}_q^{\prime\prime}(u)= \begin{cases}0, & \text { if } u \leq u_0-\varepsilon \\ 2 a\left(u-u_0\right)+b, & \text { if } u_0-\varepsilon<u<u_0+\varepsilon \\ \frac{1}{u^{q+2}}\frac{q^{q+1}}{(q+1)^q}, & \text { if } u \geq u_0+\varepsilon\end{cases}
\label{1-52}
\end{eqnarray}
Let
\begin{eqnarray}
\tilde Q(\bm\theta)=
\frac{1}{N}\sum_{m=1}^M(\bm\theta-\bm\theta^0)^T\nabla L^m(\bm\theta^0)
+\frac{1}{2N}\sum_{m=1}^M(\bm\theta-\bm\theta^0)^TH^m(\bm\theta^0)(\bm\theta-\bm\theta^0).
\label{2-3}
\end{eqnarray}
Therefore, we have
\begin{eqnarray}
\frac{\partial\tilde Q(\bm\theta)}{\partial\bm\theta}=
\frac{1}{N}\sum_{m=1}^M\nabla L^m(\bm\theta^0)
+\frac{1}{N}\sum_{m=1}^MH^m(\bm\theta^0)(\bm\theta-\bm\theta^0).
\label{2-4}
\end{eqnarray}
and
\begin{eqnarray}
\frac{\partial^2\tilde Q(\bm\theta)}{\partial\bm\theta^2}=
\frac{1}{N}\sum_{m=1}^MH^m(\bm\theta^0).
\label{2-5}
\end{eqnarray}
\\
Define 
\begin{eqnarray}
\bar{H}^m(\bm\theta^0)=\sum_{i=1}^{n^{\scriptscriptstyle(m)}}\tilde V_q''(y_i\bar{\bm{x}}_i^T\bm{\theta^0})\bar{\bm{x}}_i\bar{\bm{x}}_i^T+ n^{\scriptscriptstyle(m)}\lambda \bm I_{p+1}
\label{D-1}
\end{eqnarray}
and
\begin{eqnarray}
\bar Q(\bm\theta)=
\frac{1}{N}\sum_{m=1}^M(\bm\theta-\bm\theta^0)^T\nabla L^m(\bm\theta^0)
+\frac{1}{2N}\sum_{m=1}^M(\bm\theta-\bm\theta^0)^T\bar H^m(\bm\theta^0)(\bm\theta-\bm\theta^0).
\label{D-2}
\end{eqnarray}
$\bar{H}^m(\bm\theta^0)$ is a positive definite matrix, it follows that
$\bar{H}^m(\bm\theta^0)-H^m(\bm\theta^0)$ is nonnegative definite matrix for any 
$\bm\theta^0$, then $\tilde Q(\bm\theta)\le \bar Q(\bm\theta)$. By the MM principle,
we achieve an algorithm through minimizing $\bar Q(\bm\theta)$. That is
\begin{eqnarray}
\frac{\partial\bar Q(\bm\theta)}{\partial\bm\theta}=
\frac{1}{N}\sum_{m=1}^M\nabla L^m(\bm\theta^0)
+\frac{1}{N}\sum_{m=1}^M\bar H^m(\bm\theta^0)(\bm\theta-\bm\theta^0)=\bm 0.
\label{D-3}
\end{eqnarray}
It follows that
\begin{eqnarray}
\bm\theta=\bm\theta^0-\Big[\sum_{m=1}^M\bar H^m(\bm\theta^0)\Big]^{-1}
\sum_{m=1}^M\nabla L^m(\bm\theta^0).
\label{D-4}
\end{eqnarray}
Set $\bm\theta^t$ and $\bm\theta^{t+1}$ to be the values of step $t$ and step $t+1$, respectively, we have the iterative algorithm as follows:
\begin{eqnarray}
\bm\theta^{t+1}=\bm\theta^t-\Big[\sum_{m=1}^M\bar H^m(\bm\theta^t)\Big]^{-1}
\sum_{m=1}^M\nabla L^m(\bm\theta^t).
\label{D-5}
\end{eqnarray}

\par
\noindent
Let
\begin{eqnarray}\label{eq:1}
L_D(\bm\theta)=
\sum_{m=1}^M\sum_{i=1}^{n^{\scriptscriptstyle(m)}}V_q(y_i\bar{\bm{x}}_i^T\bm{\theta})+ \frac{n^{\scriptscriptstyle(m)}\lambda}{2}\bm\theta^T\bm W\bm\theta,\quad \hat{\bm\theta}=\underset{\bm\theta \in \mathbb{R}^{p+1}}{\arg \min}L_D(\bm\theta)\end{eqnarray}
and define
\begin{eqnarray*}
\bm\theta^*= \underset{\bm\theta \in \mathbb{R}^{p+1}}{\arg \min} \mathbb E[V_q(y\bm{x}^T\bm{\theta})+\frac{\lambda}{2}\bm{\theta}^T\bm{W}\bm{\theta}].
\end{eqnarray*}
It is easy to see that
\begin{eqnarray}\label{eq:2}
\nabla L_D(\bm\theta)=\sum_{m=1}^M\sum_{i=1}^{n^{\scriptscriptstyle(m)}}y_iV_q'(y_i\bar{\bm{x}}_i^T\bm{\theta})\bar{\bm{x}}_i+  n^{\scriptscriptstyle(m)}\lambda \bm W\bm\theta
\end{eqnarray}
and
\begin{eqnarray}\label{eq:3}
H_D(\bm\theta)=\sum_{m=1}^M\sum_{i=1}^{n^{\scriptscriptstyle(m)}}\tilde V_q''(y_i\bar{\bm{x}}_i^T\bm{\theta})\bar{\bm{x}}_i\bar{\bm{x}}_i^T+ n^{\scriptscriptstyle(m)}\lambda \bm W.
\end{eqnarray}
We further denote
\begin{eqnarray*}
\bar H_D(\bm\theta)=\sum_{m=1}^M\sum_{i=1}^{n^{\scriptscriptstyle(m)}}\tilde V_q''(y_i\bar{\bm{x}}_i^T\bm{\theta})\bar{\bm{x}}_i\bar{\bm{x}}_i^T+ n^{\scriptscriptstyle(m)}\lambda \bm I_{p+1}.
\end{eqnarray*}
\par
\noindent
According to the idea of the MM algorithm, We construct the function $u(\bm\theta,\bm\theta')$ for any $\bm\theta,\bm\theta'\in\mathbb{R}^{p+1}$ as follows,
\begin{eqnarray}\label{eq:4}
u(\bm\theta,\bm\theta')=
L_D(\bm\theta')+(\bm\theta-\bm\theta')^T\nabla L_D(\bm\theta')+\frac12(\bm\theta-\bm\theta')^T\bar H_D(\bm\theta')(\bm\theta-\bm\theta')
\end{eqnarray}
The following theorem shows that as the number of iterations increases, the estimated parameter $\boldsymbol{\theta}$ will converge to the optimal solution $\boldsymbol{\theta}^{\star}$ with high probability. 
\begin{theorem}
The iterative algorithm \eqref{D-5} converges to $\bm\theta^*$ in probability.
\end{theorem}
\noindent
\textbf{\textit{Remark 1:}}
In this algorithm \eqref{D-5}, no direct communication is required between individual clients, and each client does not need to transmit raw data to the central server. Instead, only the aggregated summary statistics, namely the gradient $\nabla L^m(\bm\theta^t)$ and the Hessian $H^m(\bm\theta^t)$, are shared. This approach not only enhances privacy protection by preventing raw data exposure but also significantly improves computational efficiency by reducing communication overhead.

\section{Online federated learning}
Many data are often received in a sequential manner rather than in a single batch. Traditional learning methods require access to the full dataset to update model parameters, which is computationally inefficient and raises significant privacy concerns. 
\\
Online federated learning is a decentralized machine learning paradigm in which multiple clients collaboratively train a model while keeping their data local. Unlike traditional batch learning approaches, online federated learning enables continuous updates to the model as new data arrives, making it well-suited for dynamic environments.
\par
We start with a sample case of two batches of data $D_1$ and $D_2$, where $D_2$ arrives after $D_1$. The idea is to update the initial $\tilde{\bm\theta}_1$(or $\tilde{\bm\theta}_1^*$) to a renewed $\tilde{\bm\theta}_2$ without using any subject level data but only some summary statistics from $D_1$.
By the equations (\ref{D-2}) and (\ref{D-3}), we have

\begin{eqnarray}
\bar Q_{D_1}(\bm\theta_1)=
\frac{1}{N_1}\sum_{m=1}^M(\bm\theta_1-\bm\theta_1^0)^T\nabla L_{D_1}^m(\bm\theta_1^0)
+\frac{1}{2N_1}\sum_{m=1}^M(\bm\theta_1-\bm\theta_1^0)^T\bar H_{D_1}^m(\bm\theta_1^0)(\bm\theta_1-\bm\theta_1^0),
\label{2-6}
\end{eqnarray}

\begin{eqnarray}
\frac{\partial\bar Q_{D_1}(\bm\theta_1)}{\partial\bm\theta_1}=
\frac{1}{N_1}\sum_{m=1}^M\nabla L_{D_1}^m(\bm\theta_1^0)
+\frac{1}{N_1}\sum_{m=1}^M\bar H_{D_1}^m(\bm\theta_1^0)(\bm\theta_1-\bm\theta_1^0).
\label{2-7}
\end{eqnarray}
and
\begin{eqnarray}
\frac{\partial^2\bar Q_{D_1}(\bm\theta_1)}{\partial\bm\theta_1^2}=
\frac{1}{N_1}\sum_{m=1}\bar H_{D_1}^m(\bm\theta_1^0).
\label{2-8}
\end{eqnarray}
Let
\begin{eqnarray}
U_1(D_1;\bm\theta_1)=
\sum_{m=1}^M\nabla L_{D_1}^m(\bm\theta_1^0)
+\sum_{m=1}^M\bar H_{D_1}^m(\bm\theta_1^0)(\bm\theta_1-\bm\theta_1^0)
\label{2-9}
\end{eqnarray}
and
\begin{eqnarray}
J_1(D_1;\bm\theta_1)=
\sum_{m=1}^M\bar H_{D_1}^m(\bm\theta_1^0).
\label{2-10}
\end{eqnarray}
$\tilde{\bm\theta}_1$ satisfies the score equation, that is
\begin{eqnarray}
\frac{1}{N_1}U_1(D_1;\tilde{\bm\theta}_1)=
\frac{1}{N_1}\sum_{m=1}^M\nabla L_{D_1}^m(\bm\theta_1^0)
+\frac{1}{N_1}\sum_{m=1}^M\bar H_{D_1}^m(\bm\theta_1^0)(\tilde{\bm\theta}_1-\bm\theta_1^0)=\bm 0.
\label{2-11}
\end{eqnarray}
In addition, $\tilde{\bm\theta}_2^*$ satisfies the following equation
\begin{eqnarray}
\frac{1}{N_2}U_1(D_1;\tilde{\bm\theta}_2^*)+ \frac{1}{N_2}U_2(D_2;\tilde{\bm\theta}_2^*)=\bm 0.
\label{2-12}
\end{eqnarray}
To achieve a renewable estimate, we take the first-order Taylor series expansion of $U_1(D_1;\tilde{\bm\theta}_2^*)$ around $\tilde{\bm\theta}_1$, we obtain
\begin{eqnarray}\nonumber
\frac{1}{N_2}U_1(D_1;\tilde{\bm\theta}_1)+ \frac{1}{N_2}J_1(D_1;\tilde{\bm\theta}_1)(\tilde{\bm\theta}_2-\tilde{\bm\theta}_1)+\frac{1}{N_2}U_2(D_2;\tilde{\bm\theta}_2)=\bm 0.
\label{2-13}
\end{eqnarray}
Since $\frac{1}{N_2}U_1(D_1;\tilde{\bm\theta}_1)=0$, 
\begin{eqnarray}\nonumber
\frac{1}{N_2}J_1(D_1;\tilde{\bm\theta}_1)(\tilde{\bm\theta}_2-\tilde{\bm\theta}_1)+\frac{1}{N_2} U_2(D_2;\tilde{\bm\theta}_2)=\bm 0.
\label{2-14}
\end{eqnarray}
It follows that
\begin{eqnarray}\nonumber
\frac{1}{N_2}J_1(D_1;\tilde{\bm\theta}_1)(\tilde{\bm\theta}_2-\tilde{\bm\theta}_1)+ \frac{1}{N_2}\sum_{m=1}^M\nabla L_{D_2}^m(\bm\theta_2^0)
+\frac{1}{N_2}\sum_{m=1}^M\bar H_{D_2}^m(\bm\theta_2^0)(\tilde{\bm\theta}_2-\bm\theta_2^0)=\bm 0.
\label{2-15}
\end{eqnarray}
It implies that
\begin{eqnarray}
[J_1(D_1;\tilde{\bm\theta}_1)+J_2(D_2;\tilde{\bm\theta}_2)]\tilde{\bm\theta}_2=
J_1(D_1;\tilde{\bm\theta}_1)\tilde{\bm\theta}_1+ J_2(D_2;\tilde{\bm\theta}_2)\bm\theta_2^0- \sum_{m=1}^M\nabla L_{D_2}^m(\bm\theta_2^0).
\label{2-16}
\end{eqnarray}
If set $\bm\theta_2^0=\tilde{\bm\theta}_1$, the proposed estimator $\tilde{\bm\theta}_2$ is 
\begin{eqnarray}
\tilde{\bm\theta}_2=
\tilde{\bm\theta}_1-[J_1(D_1;\tilde{\bm\theta}_1)+J_2(D_2;\tilde{\bm\theta}_2)]^{-1}
\sum_{m=1}^M\nabla L_{D_2}^m(\tilde{\bm\theta}_1).
\label{2-17}
\end{eqnarray}
We developed an algorithm (\ref{2-17}) based on two batches of data, $D_1$ and $D_2$, where $D_2$ arrives after $D_1$. Through this framework, we demonstrated that model parameters can be efficiently updated using only summary statistics from $D_1$, thereby preserving both privacy and computational efficiency. However, in real-world applications, data typically arrives in multiple sequential batches rather than just two. Therefore, it is essential to extend our methodology to a more general setting that accommodates continuous data updates while maintaining efficiency and privacy.
\\
In general, a renewable estimator $\tilde{\bm\theta}_b$ of $\bm\theta_b$ is defined as a solution to the following estimating equation
\begin{eqnarray}
\frac{1}{N_b}\sum_{j=1}^{b-1}J_j(D_j;\tilde{\bm\theta}_j)(\tilde{\bm\theta}_b-\tilde{\bm\theta}_{b-1})+ \frac{1}{N_b}\sum_{m=1}^M\nabla L_{D_b}^m(\bm\theta_b^0)
+\frac{1}{N_b}\sum_{m=1}^M\bar H_{D_b}^m(\bm\theta_b^0)(\tilde{\bm\theta}_b-\bm\theta_b^0)=\bm 0.
\label{2-18}
\end{eqnarray}
Equivalently, 
\begin{equation}
\begin{aligned}
\tilde{\boldsymbol{\theta}}_b &= \arg \min _{\boldsymbol{\theta}}\bigg\{\frac{1}{2N_b}(\boldsymbol{\theta}-\tilde{\boldsymbol{\theta}}_{b-1})^T\sum_{j=1}^{b-1}J_j(D_j;\tilde{\boldsymbol{\theta}}_j)(\boldsymbol{\theta}-\tilde{\boldsymbol{\theta}}_{b-1}) 
+ \frac{1}{N_b}\sum_{m=1}^M(\boldsymbol{\theta}-\tilde{\boldsymbol{\theta}}_{b-1}) ^T\nabla L_{D_b}^m(\boldsymbol{\theta}_b^0)\\
& \quad +\frac{1}{2N_b}(\boldsymbol{\theta}-\boldsymbol{\theta}_b^0)^T\sum_{m=1}^M\bar H_{D_b}^m(\boldsymbol{\theta}_b^0)(\boldsymbol{\theta}-\boldsymbol{\theta}_b^0)\bigg\}.
\end{aligned}
\label{2-18-1}
\end{equation}
In addition, (\ref{2-18}) yields the following result
\begin{eqnarray}
[\sum_{j=1}^{b-1}J_j(D_j;\tilde{\bm\theta}_j)+J_b(D_b;\tilde{\bm\theta}_b)]\tilde{\bm\theta}_b
=
\sum_{j=1}^{b-1}J_1(D_j;\tilde{\bm\theta}_j)\tilde{\bm\theta}_{b-1}+ J_b(D_b;\tilde{\bm\theta}_b)\bm\theta_b^0- \sum_{m=1}^M\nabla L_{D_b}^m(\bm\theta_b^0).
\label{2-19}
\end{eqnarray}
Similarly, if set $\bm\theta_b^0=\tilde{\bm\theta}_{b-1}$, the proposed estimator $\tilde{\bm\theta}_b$ is 
\begin{eqnarray}
\tilde{\bm\theta}_b=
\tilde{\bm\theta}_{b-1}-[\sum_{j=1}^{b}J_j(D_j;\tilde{\bm\theta}_j)]^{-1}
\sum_{m=1}^M\nabla L_{D_b}^m(\tilde{\bm\theta}_{b-1}),
\label{2-20}
\end{eqnarray}
where
\begin{eqnarray}
J_j(D_j;\tilde{\bm\theta}_j)=
\sum_{m=1}^M\bar H_{D_j}^m(\bm\theta_j^0),
\label{2-21}
\end{eqnarray}
\begin{equation}
\begin{aligned}
\bar H_{D_j}^m(\bm\theta_j^0)&=
\sum_{i=1}^{n_j^{\scriptscriptstyle(m)}}\tilde V_q^{\prime\prime}(y_{{\scriptscriptstyle D_{\! j}}i}\bar{\bm{x}}_{{\scriptscriptstyle D_{\! j}}i}^T\bm{\theta}_j^0)
\bar{\bm{x}}_{{\scriptscriptstyle D_{\! j}}i}\bar{\bm{x}}_{{\scriptscriptstyle D_{\! j}}i}^T
+ n_j^{\scriptscriptstyle(m)}\lambda_j\bm I_{p+1}\\
&=\sum_{i=1}^{n_j^{\scriptscriptstyle(m)}}\tilde V_q^{\prime\prime}(y_{{\scriptscriptstyle D_{\! j}}i}\bar{\bm{x}}_{{\scriptscriptstyle D_{\! j}}i}^T\tilde{\bm\theta}_{j-1})
\bar{\bm{x}}_{{\scriptscriptstyle D_{\! j}}i}\bar{\bm{x}}_{{\scriptscriptstyle D_{\! j}}i}^T
+n_j^{\scriptscriptstyle(m)}\lambda_j\bm I_{p+1}.
\end{aligned}
\label{2-22}
\end{equation}
and
\begin{eqnarray}
\nabla L_{D_b}^m(\tilde{\bm\theta}_{b-1})=
\sum_{i=1}^{n_b^{\scriptscriptstyle(m)}}y_{{\scriptscriptstyle D_{\! b}}i}V_q^{\prime}(y_{{\scriptscriptstyle D_{\! b}}i}\bar{\bm{x}}_{{\scriptscriptstyle D_{\! d}}i}^T\tilde{\bm\theta}_{b-1})
\bar{\bm{x}}_{{\scriptscriptstyle D_{\! b}}i}+n_b^{\scriptscriptstyle(m)}\lambda_b\bm W\tilde{\bm\theta}_{b-1}.
\label{2-23}
\end{eqnarray}
That is, 
the proposed estimator $\tilde{\bm\theta}_b$ is 
\begin{eqnarray}
\tilde{\bm\theta}_b=
\tilde{\bm\theta}_{b-1}-\Big[\sum_{j=1}^{b}\sum_{m=1}^M\bar H_{D_j}^m(\tilde{\bm\theta}_{j-1})\Big]^{-1}
\sum_{m=1}^M\nabla L_{D_b}^m(\tilde{\bm\theta}_{b-1}),
\label{2-20}
\end{eqnarray}
where
\begin{eqnarray}
\bar H_{D_j}^m(\tilde{\bm\theta}_{j-1})=
\sum_{i=1}^{n_j^{\scriptscriptstyle(m)}}\tilde V_q^{\prime\prime}(y_{{\scriptscriptstyle D_{\! j}}i}(1,\bm{x}_{{\scriptscriptstyle D_{\! j}}i}^T)\tilde{\bm\theta}_{j-1})
(1,\bm{x}_{{\scriptscriptstyle D_{\! j}}i}^T)^T(1,\bm{x}_{{\scriptscriptstyle D_{\! j}}i}^T)+n_j^{\scriptscriptstyle(m)}\lambda_j\bm I_{p+1}
\label{2-22}
\end{eqnarray}
and
\begin{eqnarray}
\nabla L_{D_b}^m(\tilde{\bm\theta}_{b-1})=
\sum_{i=1}^{n_b^{\scriptscriptstyle(m)}}y_{{\scriptscriptstyle D_{\! b}}i}V_q^{\prime}(y_{{\scriptscriptstyle D_{\! b}}i}(1,\bm{x}_{{\scriptscriptstyle D_{\! b}}i}^T)\tilde{\bm\theta}_{b-1})
(1,\bm{x}_{{\scriptscriptstyle D_{\! b}}i}^T)^T+n_b^{\scriptscriptstyle(m)}\lambda_b\bm W\tilde{\bm\theta}_{b-1}.
\label{2-23}
\end{eqnarray}

\begin{table*}[]
{\begin{tabular}[l]{@{}lccccccccc}
\toprule
{\bf{Algorithm 1:}} Online federated learning algorithm for classification\\[1ex]
\midrule
 1. 
Input: streaming data sets $D_1, \ldots, D_b, \ldots$ and parameters $\lambda$, $q$
\\
2. Initialize: set initial value
$\tilde{\bm{\theta}}_0$
\\
3. For $b=1,2,\ldots$ do\\
4. Read in data set $D_b$
\\
5. Compute $H_{D_b}^m(\tilde{\bm\theta}_{b-1})$ and $\nabla L_{D_b}^m(\tilde{\bm\theta}_{b-1})$ on each machine independently $m=1,2, \ldots, M$, \\ ~~~~then send them to the central server
\\
6. Update $\tilde{\bm\theta}_b=
\tilde{\bm\theta}_{b-1}-\Big[\sum_{j=1}^{b}\sum_{m=1}^MH_{D_j}^m(\tilde{\bm\theta}_{j-1})\Big]^{-1}
\sum_{m=1}^M\nabla L_{D_b}^m(\tilde{\bm\theta}_{b-1})$ on the central server
\\
7. End
\\
8. Output: sign$(\bar{\bm{x}}^T\tilde{\bm\theta}_b)$.\\
\bottomrule
\end{tabular}}\\
\end{table*}
\par\noindent
\textbf{\textit{Remark 2:}}
Algorithm 1 provides an efficient and scalable approach for federated learning in an online setting. Unlike traditional batch learning methods, which require retraining on the entire dataset when new data arrives, our algorithm updates the model incrementally using only summary statistics. This significantly reduces computational overhead and memory usage, making it well-suited for large-scale, real-time applications. Furthermore, since individual clients only share aggregated statistics with the central server, Algorithm 1 enhances privacy protection by preventing direct data transmission, aligning with federated learning principles.

\section{Online federated learning with differential privacy}

In many real-world applications, FL involves sensitive user data, making privacy preservation a crucial component of any practical implementation. 
To overcome these challenges, differential privacy has emerged as a powerful technique for providing formal privacy guarantees. Recent studies have successfully integrated DP into federated learning systems and enabled robust privacy protection while maintaining model performance.
Building upon this foundation, this section extends our previous online learning approach by incorporating privacy-preserving mechanisms. We define the necessary privacy guarantees, introduce algorithmic modifications to ensure compliance with differential privacy constraints, and analyze the theoretical implications of these modifications on model performance and security.

\noindent{\bf Definition 1} (Differential privacy \cite{dwork2006calibrating}). {\sl 
An algorithm \( \mathcal{A} \) satisfies \((\epsilon, \delta)\)-differential privacy if, for any two datasets \( D \) and \( D' \) differing by at most one element, and for all Borel sets $O \subseteq \operatorname{Range}(\mathcal{A})$,

\[
\Pr[\mathcal{A}(D) \in O] \leq e^\epsilon \Pr[\mathcal{A}(D') \in O] + \delta,
\]
where
\( \epsilon \geq 0 \) is the \textit{privacy budget} that controls the trade-off between privacy and accuracy. Smaller \( \epsilon \) means stronger privacy.
\( \delta \geq 0 \) is the probability of allowing a small relaxation of the strict privacy guarantee.
\( D \) and \( D' \) are neighboring datasets, differing by only one record.
For \( \delta = 0 \), the definition reduces to pure \( \epsilon \)-differential privacy.
}
\\
Define
\begin{equation}
\begin{aligned}
G(\bm\theta)
&=l_{D_b}(\tilde{\boldsymbol{\theta}}_{b-1})+ 
\frac{1}{N_b}(\boldsymbol{\theta}-\tilde{\boldsymbol{\theta}}_{b-1})^T\nabla L_{D_b}(\tilde{\bm\theta}_{b-1})+
\frac{1}{2N_b}(\boldsymbol{\theta}-\tilde{\boldsymbol{\theta}}_{b-1})^T\sum_{j=1}^{b}\bar H_{D_j}(\tilde{\boldsymbol{\theta}}_{j-1})(\boldsymbol{\theta}-\tilde{\boldsymbol{\theta}}_{b-1})
\end{aligned}
\label{G-1}
\end{equation}
and
\begin{equation}
\begin{aligned}
Q(\bm\theta)=
G(\bm\theta)+\frac{\rho}{2N_b}\|\bm\theta\|^2+\frac{\bm\xi^T\bm\theta}{N_b},
\end{aligned}
\label{Q-1}
\end{equation}
where $ \nabla L_{D_b}(\tilde{\bm\theta}_{b-1})=
\sum_{m=1}^M\nabla L_{D_b}^m(\tilde{\bm\theta}_{b-1})$, $ \bar H_{D_j}(\tilde{\bm\theta}_{j-1})=
\sum_{m=1}^M \bar H_{D_j}^m(\tilde{\bm\theta}_{j-1})$, $\rho$ is positive constant.
\\
In this work, we assume that the random vector $\bm\xi \in \mathbb{R}^{p+1}$ is drawn from two different distributions.
\\
$\bm\xi$ is drawn from Laplace distribution with Laplace density function 
\begin{equation}
\begin{aligned}
p(\bm\xi)=\frac{1}{(2\eta)^{p+1}} e^{-\|\bm\xi\|_1 /\eta},
\end{aligned}
\label{La-1}
\end{equation}
where $\|\bm\xi\|_1$ denotes the $L_1$ norm of $\bm\xi$, $\eta$ is a positive constant.
\\
$\bm\xi$ is drawn from $\mathcal{N}\left(\mathbf{0}, \tau^2 \boldsymbol{I}_{p+1}\right)$ with density function 
\begin{equation}
\begin{aligned}
f(\boldsymbol{\xi})=\frac{1}{(2 \pi\tau^2)^{\frac{p+1}{2}}} \exp \left(-\frac{1}{2 \tau^2}\|\boldsymbol{\xi}\|^2\right).
\end{aligned}
\label{La-2}
\end{equation}
\\
We achieve the proposed estimator with privacy preservation by minimizing the following objective function. That is
\begin{equation}
\begin{aligned}
 \tilde{\boldsymbol{\theta}}_b^p=\arg \min _{\boldsymbol{\theta}}Q(\bm\theta).
\end{aligned}
\label{Q-2}
\end{equation}
Furthermore, the proposed estimator with privacy preservation is expressed as
\begin{equation}
\begin{aligned}
\tilde{\bm\theta}_b^{p}=
\left[\sum_{j=1}^{b}J_j(D_j;\tilde{\bm\theta}_j^p)+\rho\bm I_{p+1}\right]^{-1}\left[\sum_{j=1}^{b}J_j(D_j;\tilde{\bm\theta}_j^p)\tilde{\bm\theta}_{b-1}^p-\nabla L_{D_b}(\tilde{\bm\theta}_{b-1}^p)-\bm\xi\right].
\end{aligned}
\label{Q-3}
\end{equation}

Let $\bar{D}_b=\left\{\left(\boldsymbol{x}_1, y_1\right), \cdots,\left(\boldsymbol{x}_{n_b-1}, y_{n_b-1}\right)\right\}, d_b=\left(\boldsymbol{x}_{n_b}, y_{n_b}\right)$ and $d_b^{\prime}=\left(\boldsymbol{x}_{n_b}^{\prime}, y_{n_b}^{\prime}\right)$. Assuming that $D_b$ and $D_b^{\prime}$ have a different value in the $n$-th item, we have $D_b=\left(\bar{D}_b ; d_b\right)$ and $D_b^{\prime}=\left(\bar{D}_b ; d_b^{\prime}\right)$.
The following regularity conditions are postulated.
\par
\textit{Condition 1}. $\left\|\overline{\boldsymbol{x}}_{n_b}\right\|_1 \leq C_1$, $\left\|\overline{\boldsymbol{x}}_{n_b}^{\prime}\right\|_1 \leq C_1,\left\|\overline{\boldsymbol{x}}_{n_b}\right\| \leq C_2,\left\|\overline{\boldsymbol{x}}_{n_b}^{\prime}\right\| \leq C_2$.
\par
\textit{Condition 2}.
 $\rho\ge \frac{(q+1)^2C_2^2}{(e^{\frac{\epsilon}{4}}-1)q}-N_b\lambda$.

\begin{table*}[]
{\begin{tabular}[l]{@{}lccccccccc}
\toprule
{\bf{Algorithm 2:}} Online federated learning with differential privacy algorithm for classification\\[1ex]
\midrule
 1. 
Input: streaming data sets $D_1, \ldots, D_b, \ldots$ and parameters $\lambda$, $q$ and $\rho$
\\
2. Initialize: set initial value
$\tilde{\bm{\theta}}_0$
\\
3. For $b=1,2,\ldots$ do\\
4. Read in data set $D_b$
\\
5. Compute $H_{D_b}^m(\tilde{\bm\theta}_{b-1})$ and $\nabla L_{D_b}^m(\tilde{\bm\theta}_{b-1})$ on each machine independently $m=1,2, \ldots, M$, \\ ~~~~then send them to the central server
\\
6. If $\boldsymbol{\xi}$ is drawn from Laplace distribution $p(\boldsymbol{\xi})=\frac{1}{(2 \eta)^{p+1}} e^{-\|\boldsymbol{\xi}\|_1 / \eta}$ with $\eta=\left(\epsilon-T_2\right)^{-1} T_1$,\\~~~ then $\epsilon$-differential privacy is satisfied, where $T_1$ and $T_2$ in (\ref{alproof-11-1})
\\
7. If $\boldsymbol{\xi}$ is drawn from $\mathcal{N}\left(\mathbf{0}, \tau^2 \boldsymbol{I}_{p+1}\right)$ with 
$\tau=\Delta_1\left(\sqrt{2 \ln \frac{1}{\delta}}+\sqrt{2 \ln \frac{1}{\delta}+\epsilon}\right) \epsilon^{-1}$,\\~~~
then $(\epsilon, \delta)$-differential privacy is satisfied, where $\Delta_1$ in (\ref{alproof-14})
\\
8. Update $\tilde{\boldsymbol{\theta}}_b^p=\left[\sum_{j=1}^b J_j\left(D_j ; \tilde{\boldsymbol{\theta}}_j^p\right)+\rho \boldsymbol{I}_{p+1}\right]^{-1}\left[\sum_{j=1}^b J_j\left(D_j ; \tilde{\boldsymbol{\theta}}_j^p\right) \tilde{\boldsymbol{\theta}}_{b-1}^p-\nabla L_{D_b}\left(\tilde{\boldsymbol{\theta}}_{b-1}^p\right)-\boldsymbol{\xi}\right]$\\
~~~ on the central server
\\
9. End
\\
10. Output: sign$(\bar{\bm{x}}^T\tilde{\bm\theta}_b^p)$.\\
\bottomrule
\end{tabular}}\\
\end{table*}
\par
\noindent
\textbf{\textit{Remark 3:}}
Algorithm 2 extends Algorithm 1 by incorporating differential privacy, ensuring that the model updates remain privacy-preserving. While Algorithm 1 already enhances privacy by avoiding direct data sharing between clients and the central server, Algorithm 2 further strengthens privacy guarantees by introducing noise into the optimization process. Specifically, Algorithm 2 perturbs the model parameters using Laplace or Gaussian noise, satisfying $\epsilon$-differential privacy or $(\epsilon, \delta)$-differential privacy, respectively. This added privacy protection comes at a slight cost in accuracy due to the noise but is essential for applications where strong privacy constraints are required. In summary, Algorithm 1 prioritizes computational efficiency and scalability, while Algorithm 2 balances efficiency with formal privacy guarantees, making it suitable for privacy-sensitive federated learning environments.

\noindent{\bf Lemma 1} \citep{chaudhuri2011differentially}. {\sl 
If $\bm P$ is a full-rank matrix and if $\bm Q$ is a matrix with rank at most 2, then,
\begin{equation}
\begin{aligned}
\frac{\operatorname{det}(\bm {P}+\bm {Q})-\operatorname{det}(\bm P)}{\operatorname{det}(\bm P)}=\gamma_1\left(\bm P^{-1} \bm Q\right)+\gamma_2\left(\bm P^{-1} \bm Q\right)+\gamma_1\left(\bm P^{-1} \bm Q\right) \gamma_2\left(\bm P^{-1} \bm Q\right),
\end{aligned}
\label{lemma-1}
\end{equation}
where $\gamma_i(\bm Z)$ is the $i$-th highest eigenvalue of matrix $\bm Z$.
}
\par
The following theorem establishes that Algorithm 2 ensures differential privacy under two different noise injection mechanisms.
\par
\noindent{\bf Theorem 2.} { Under conditions 1 and 2, Algorithm 2 is $\epsilon$-differentially private if the random vector $\bm\xi$ follows Laplace distribution with density function in (\ref{La-1}) and is $(\epsilon, \delta)$-differentially private if the random vector $\bm\xi$ follows normal distribution with density function in (\ref{La-2}). 
}
\par
The propositions 1 and 2 below demonstrate that the proposed estimator $\tilde{\bm\theta}_{b}^p$ in (\ref{Q-3}) achieves both consistency and asymptotic normality.
\par
\noindent{\bf Proposition 1.} {\sl  The estimator given in (\ref{Q-3}) is consistent, that is $\tilde{\bm\theta}_{b}^p\xrightarrow{\mathrm{P}}\bm\theta_0$, as $N_b\rightarrow\infty$}.
\noindent
\textbf{\textit{Note 1:}}
Proposition 1 establishes the consistency of the proposed estimator $\tilde{\boldsymbol{\theta}}_b^p$, meaning that as the number of observations $N_b$ increases, the estimator converges in probability to the true parameter $\boldsymbol{\theta}_0$. This guarantees that our method provides reliable parameter estimates given sufficient data. The consistency result is fundamental because it ensures that the estimator remains stable and accurate as more data becomes available in the federated online learning setting.
\par
\noindent{\bf Proposition 2} {\sl  (Asymptotic Normality). Suppose $\bm\theta_0$ =O(1). 
The proposed estimator $\tilde{\boldsymbol{\theta}}_b^p$ follows an asymptotic normal distribution. That is, as the total sample size $N_b$ goes to infinity,
\begin{eqnarray}
\sqrt{N_b}\left(\tilde{\boldsymbol{\theta}}_b^p-\boldsymbol{\theta}_0\right) \xrightarrow{\mathrm{D}} \mathcal{N}\left(\bm 0, \boldsymbol{\Sigma}_0\right),
\label{th3-1}
\end{eqnarray}
}
where $\boldsymbol{\Sigma}_0$  represents the inverse of the Fisher information for a single observation at the true value. 
\par
\noindent
\textbf{\textit{Note 2:}}
Proposition 2 further strengthens the theoretical foundation of our method by proving the asymptotic normality of $\tilde{\boldsymbol{\theta}}_b^p$. This result implies that, for large sample sizes, the estimator follows a normal distribution centered around the true parameter $\boldsymbol{\theta}_0$ with variance given by the inverse Fisher information matrix. 
\par
\noindent{\bf Lemma 2} \citep{wang2018another}. {Let $f_{\bm\theta}(\bm{X})=\bar{\bm{X}}^T\bm{\theta}$, $R(\bm{\theta})=E_{\bm{X}Y}[Y\neq$ sign$(f_{\bm\theta}(\bm{X}))]$ and $\bm\theta^*=\arg \min _{\bm\theta} R(\bm\theta)$, then
\begin{eqnarray}
R(\tilde{\bm\theta}_b^p)-R(\bm\theta^*)\leqslant \frac{q+1}{q}\varepsilon_E,
\label{lem2-1}
\end{eqnarray}
where $\varepsilon_{E}$ is defined as follows and $V_q$ is the generalized DWD loss:
$$
\begin{gathered}
\varepsilon_{E}=E_{\bm{X} Y}[V_q\{Y f_{\tilde{\bm\theta}_b^p}(\bm X)\}]-E_{\bm X Y}[V_q\{Y f_{\bm\theta_0}(\bm X)\}] .
\end{gathered}
$$
}

The following theorem shows that the proposed online classifier with privacy preservation achieves Bayesian risk consistency.
\par
\noindent{\bf Theorem 3.} {\sl Suppose $\lambda=O(1)$ and $\bm\theta_0=O(1)$, then
$R(\tilde{\bm\theta}_b^p)\xrightarrow{\mathrm{P}} R(\bm\theta^*)$.}
\par
\noindent
\textbf{\textit{Remark 4:}}
Theorem 3 establishes that as the sample size increases, the expected misclassification rate of the classifier, $R(\tilde{\boldsymbol{\theta}}_b^p)$, converges to the optimal Bayes risk $R(\boldsymbol{\theta}^*)$, which represents the minimum possible classification error under the true data distribution.
This result is crucial for practical applications because it guarantees that our method remains effective even as new data arrives, ensuring that privacy preservation does not significantly degrade classification performance. Furthermore, the result highlights that despite the introduction of privacy-preserving noise in Algorithm 2, the model maintains strong generalization capabilities, making it robust for real world federated learning scenarios where data privacy and adaptability are both critical.

\section{Simulation}
In this section, we compare the online generalized DWD with privacy-preserving features against the offline generalized DWD with privacy-preserving features and the existing generalized DWD method \citep{wang2018another} in terms of classification accuracy and computational speed.
The positive class is sampled from $\mathcal{N}_p\left(\mu\boldsymbol{1}_p, \sigma^2 \boldsymbol{I}_p\right)$, while the negative class is drawn from $\mathcal{N}_p\left(-\mu\boldsymbol{1}_p, \sigma^2 \boldsymbol{I}_p\right)$, where $\boldsymbol{1}_p$ denotes a $p$-dimensional column vector with all elements equal to 1 and $\boldsymbol{I}_p$ is the $p \times p$ identity matrix. In this simulation, we consider two scenarios for the parameters $\mu$ and $\sigma$. In the first scenario, the parameters $\mu$ and $\sigma$ are consistent across all sites. In the second scenario, $\mu$ and $\sigma$ vary between sites and are randomly generated. The data generated includes both `balanced' and `imbalanced' datasets. Balanced data refers to cases where the ratio of the positive class to the negative class is 1:1, while imbalanced data indicates unequal quantities between the two classes.
\par
We evaluate the proposed method's prediction accuracy, stability, and time cost by varying different parameter values. In the following discussion, `OnWDP' refers to the online federated learning with differential privacy algorithm for DWD, `OnWP' refers to the online federated learning without differential privacy algorithm for DWD, `OffWP' stands for the offline federated learning without differential privacy algorithm for DWD, and `OffNP' represents the generalized DWD method. The generalized DWD method is implemented using the R package \texttt{kerndwd} with the kernel parameter set to \texttt{kern = vanilladot()}.

\begin{figure}[htbp]
    \centering
    \begin{subfigure}[b]{0.48\textwidth}
        \centering
        \includegraphics[width=\textwidth]{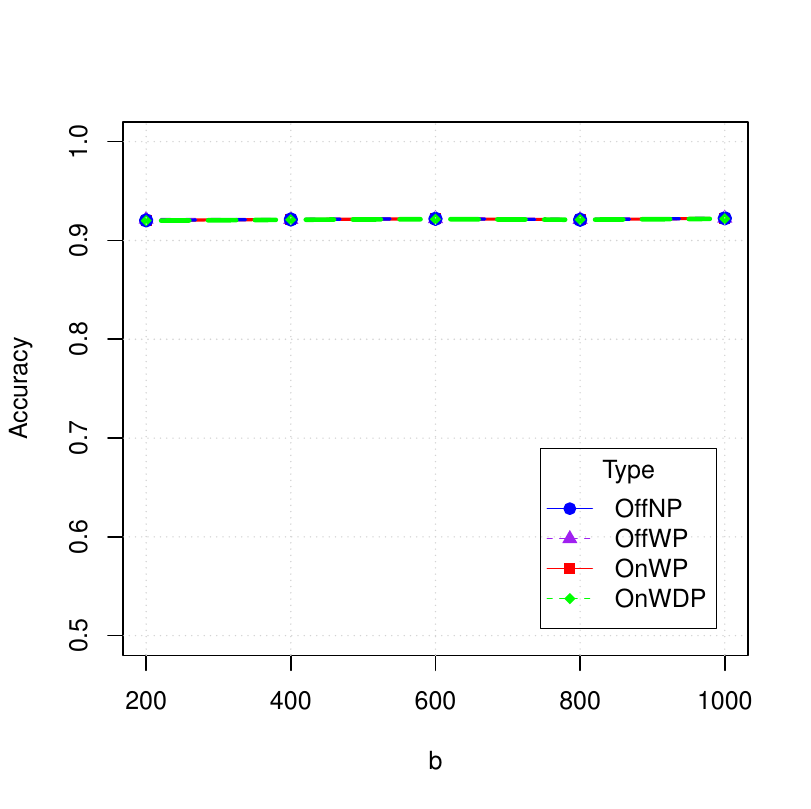}
        \caption{
}
        \label{fig:sub1}
    \end{subfigure}
    \hfill
    \begin{subfigure}[b]{0.48\textwidth}
        \centering
        \includegraphics[width=\textwidth]{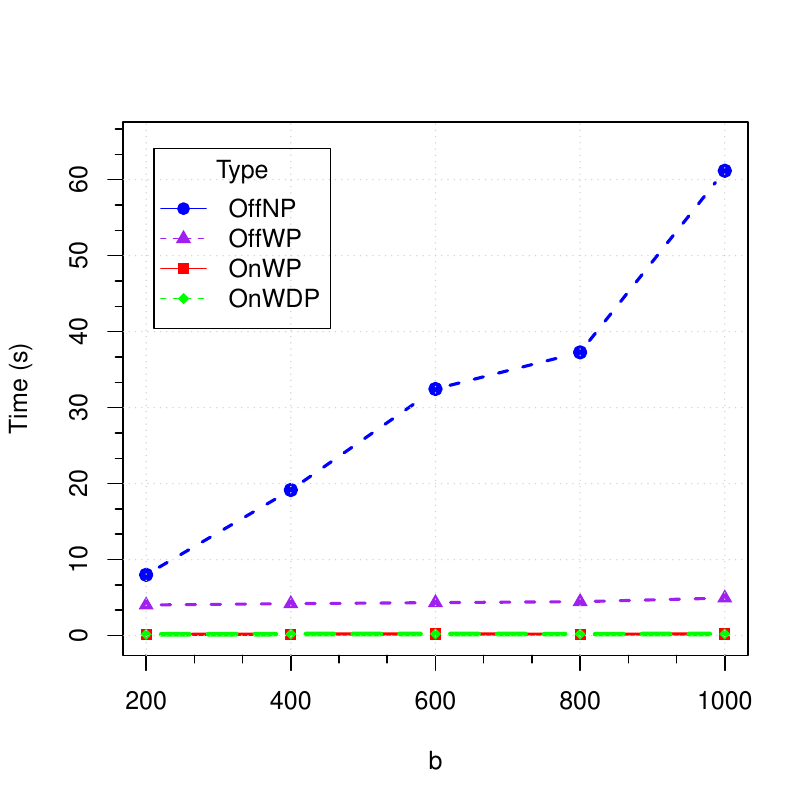}
        \caption{}
        \label{fig:sub2}
    \end{subfigure}
    \vfill
    \caption{Comparison of accuracy and time cost for the balanced dataset (two-class data, 1:1 ratio) based on variations in parameter $b$, with $M$=20, $p$=50, $\mu$=0.2, $\sigma$=1, $q$=1, $\epsilon=0.8$ and $\delta=10^{-5}$.}
    \label{fig2}
\end{figure}

\begin{figure}[htbp]
    \centering
    \begin{subfigure}[b]{0.48\textwidth}
        \centering
        \includegraphics[width=\textwidth]{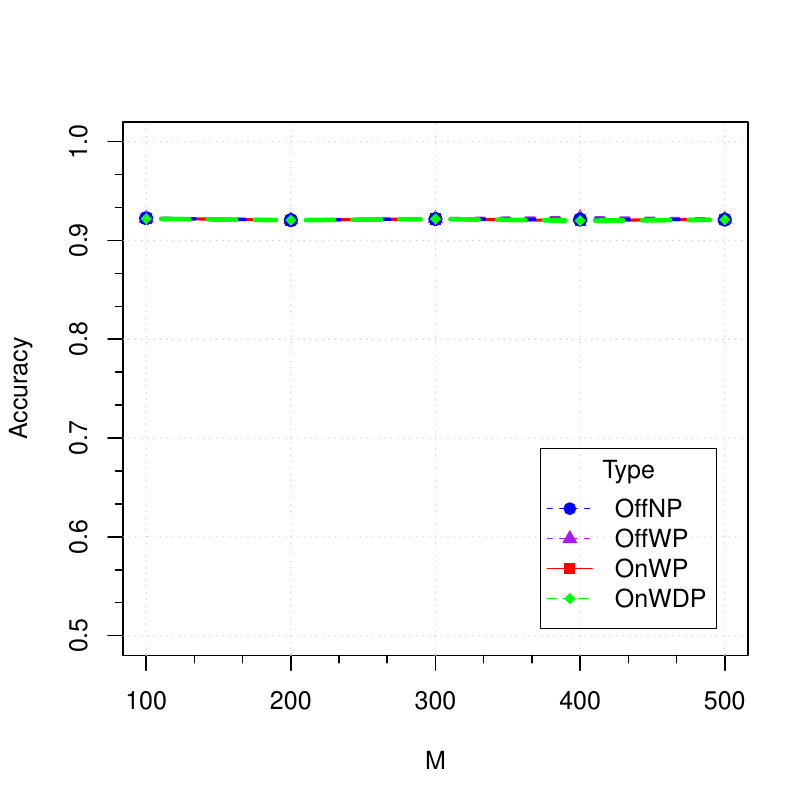}
        \caption{
}
        \label{fig:sub1}
    \end{subfigure}
    \hfill
    \begin{subfigure}[b]{0.48\textwidth}
        \centering
        \includegraphics[width=\textwidth]{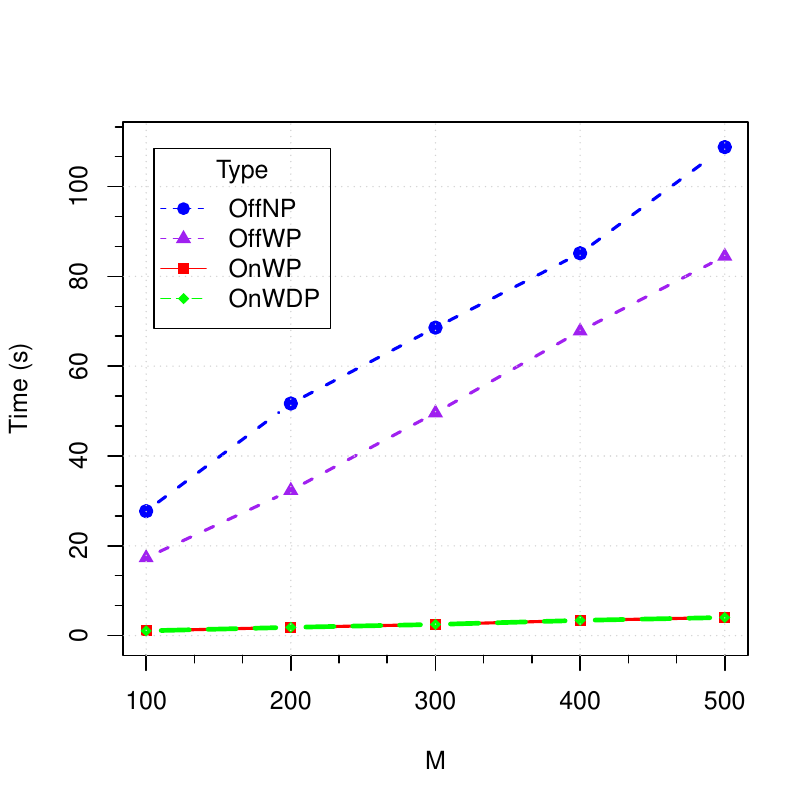}
        \caption{}
        \label{fig:sub2}
    \end{subfigure}
    \vfill
       \caption{Comparison of accuracy and time cost for the balanced dataset (two-class data, 1:1 ratio) based on variations in parameter $M$, with $b$=100, $p$=50, $\mu$=0.2, $\sigma$=1, $q$=1, $\epsilon=0.8$ and $\delta=10^{-5}$.}
    \label{fig3}
\end{figure}

\begin{figure}[htbp]
    \centering
    \begin{subfigure}[b]{0.48\textwidth}
        \centering
        \includegraphics[width=\textwidth]{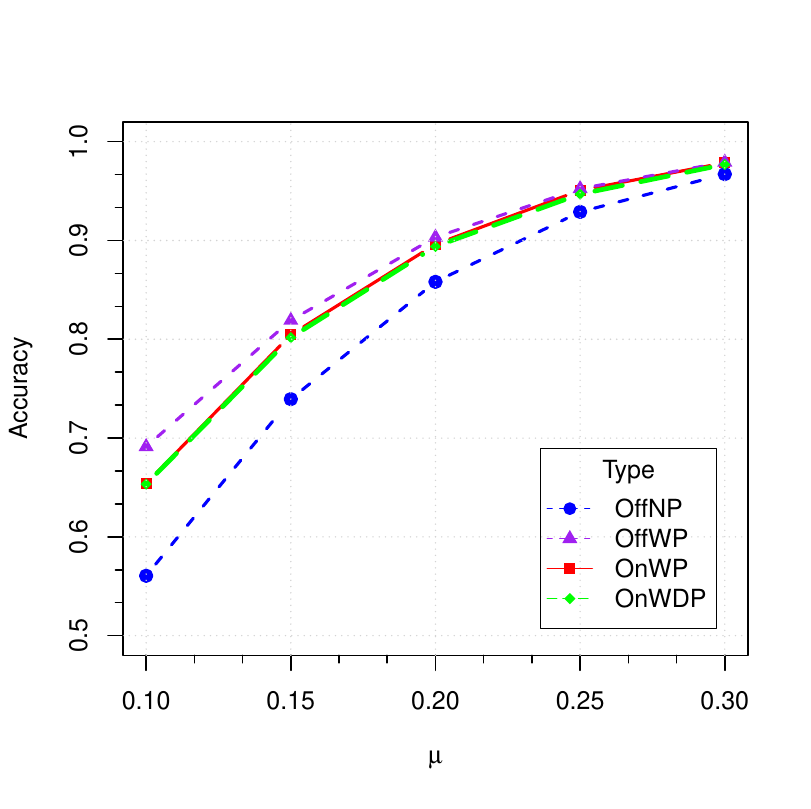}
        \caption{
}
        \label{fig:sub1}
    \end{subfigure}
    \hfill
    \begin{subfigure}[b]{0.48\textwidth}
        \centering
        \includegraphics[width=\textwidth]{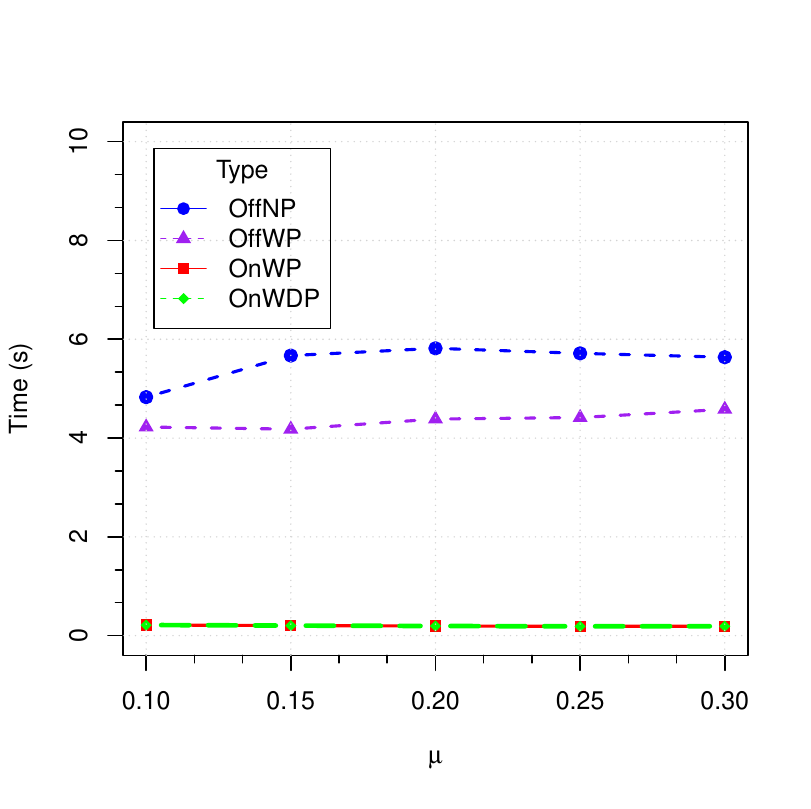}
        \caption{}
        \label{fig:sub2}
    \end{subfigure}
    \vfill
     \caption{Comparison of accuracy and time cost for the imbalanced dataset (two-class data, 4:1 ratio) based on variations in parameter $\mu$, with $M$=20, $p$=50, $b$=100, $\sigma$=1, $q$=1, $\epsilon=0.8$ and $\delta=10^{-5}$.}
    \label{fig4}
\end{figure}

\begin{figure}[htbp]
    \centering
    \begin{subfigure}[b]{0.48\textwidth}
        \centering
        \includegraphics[width=\textwidth]{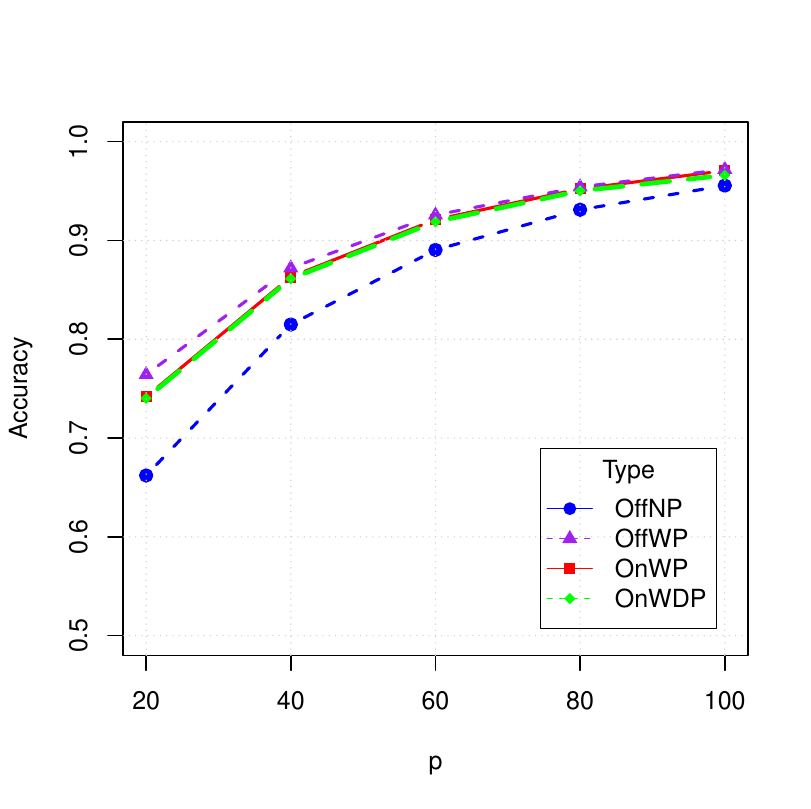}
        \caption{
}
        \label{fig:sub1}
    \end{subfigure}
    \hfill
    \begin{subfigure}[b]{0.48\textwidth}
        \centering
        \includegraphics[width=\textwidth]{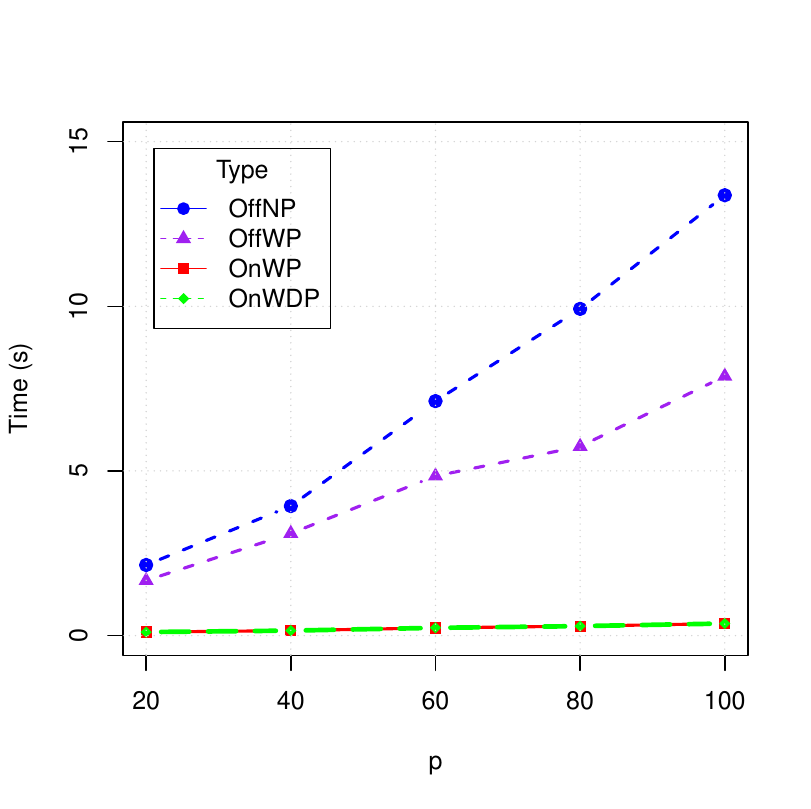}
        \caption{}
        \label{fig:sub2}
    \end{subfigure}
    \vfill
   \caption{Comparison of accuracy and time cost for the imbalanced dataset (two-class data, 4:1 ratio) based on variations in parameter $p$, with $M$=20, $\mu$=0.2, $b$=100, $\sigma$=1, $q$=1, $\epsilon=0.8$ and $\delta=10^{-5}$.}
    \label{fig5}
\end{figure}

\begin{figure}[htbp]
    \centering
    \begin{subfigure}[b]{0.48\textwidth}
        \centering
        \includegraphics[width=\textwidth]{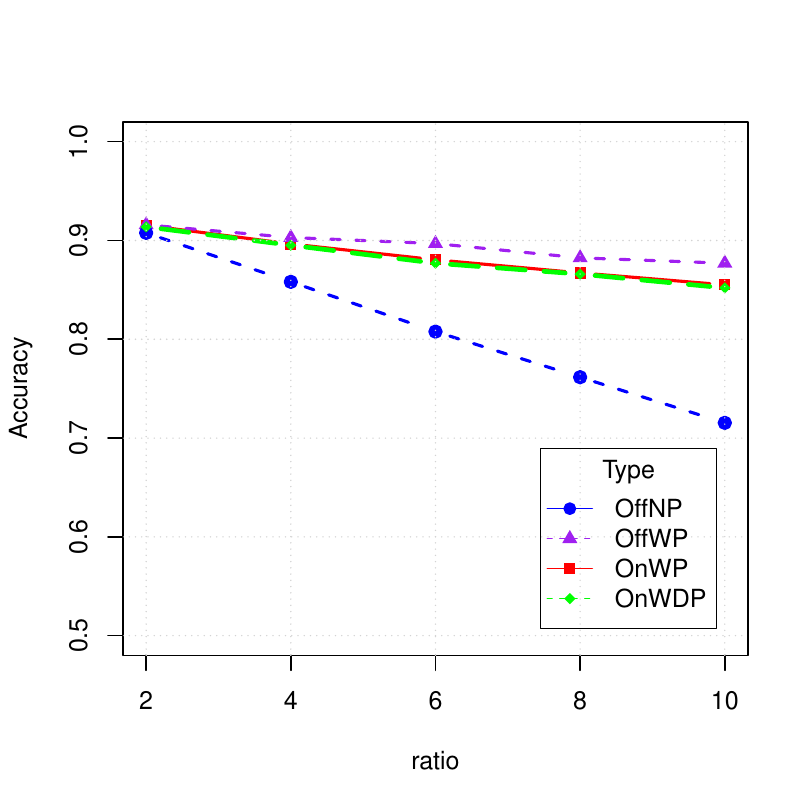}
        \caption{
}
        \label{fig:sub1}
    \end{subfigure}
    \hfill
    \begin{subfigure}[b]{0.48\textwidth}
        \centering
        \includegraphics[width=\textwidth]{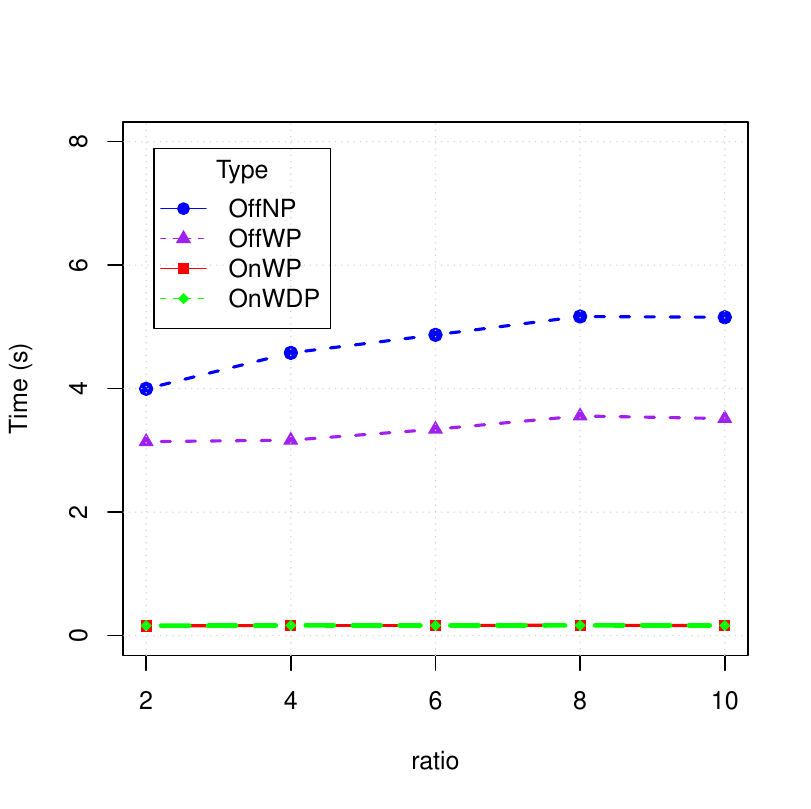}
        \caption{}
        \label{fig:sub2}
    \end{subfigure}
    \vfill
    \caption{Comparison of accuracy and time cost for the imbalanced dataset based on variations in the two-class data ratio, with $M$=20, $p$=50, $b$=100, $\mu$=0.2, $\sigma$=1, and $q$=1. `ratio' indicates that the ratio of the positive class to the negative class.}
    \label{fig6}
\end{figure}

The Tables \ref{Tab1}-\ref{Tab3} and Figures \ref{fig2}-\ref{fig6} provide a comprehensive comparison of four methods, OffNP, OffWP, OnWP and OnWDP, in terms of accuracy and computation speed, especially across balanced and imbalanced datasets. 
\\
Accuracy comparison:
For balanced datasets, all four methods perform similarly, showing only minor differences. The OffWP and OnWP methods slightly outperform OnWDP and OffNP in some cases.
The performance differences become more noticeable in imbalanced datasets, where the proposed online privacy-preserving methods (OnWDP and OnWP) show clear superiority over OffNP, the OffWP method is also better than OffNP. In these scenarios, OffNP's accuracy ranges between 85-90\%, while OnWDP, OnWP and OffWP consistently deliver higher accuracies, depending on the dataset size and feature dimension. For instance, in smaller imbalanced datasets ($M=10, b=100, p=50$), OffNP's accuracy is around 85 \%, but OnWDP, OnWP and OffWP achieve over 88\%, demonstrating their enhanced ability to handle skewed class distributions. This trend continues across larger datasets and various parameter configurations, such as changes in the feature dimension ($p$), the number of data batches ($b$), or the number of sites ($M$).
\\
Computation time comparison:
Computational speed is where the most substantial differences are observed. OffNP is notably slower than  OnWDP, OnWP and OffWP across all dataset sizes and configurations.  OnWDP and OnWP consistently outperform the other two methods in terms of computational speed, regardless of dataset size or feature dimension.
As the dataset size grows (e.g., $M=10, b=2000, p=50$ ), the computation time for OffNP increases drastically to over 50 seconds, while OffWP handles the same workload in approximately 10 seconds. OnWDP and OnWP, in contrast, continue to complete the task in around 0.1 seconds, demonstrating its scalability and computational efficiency. Even with larger feature dimensions or more data points (as shown in Table \ref{Tab2}), OnWDP and OnWP consistently outperform both OffWP and OffNP, with near-instant computation times, while OffNP may take significantly longer.
\par
\begin{figure}[htbp]
    \centering
    \begin{subfigure}[b]{0.48\textwidth}
        \centering
        \includegraphics[width=\textwidth]{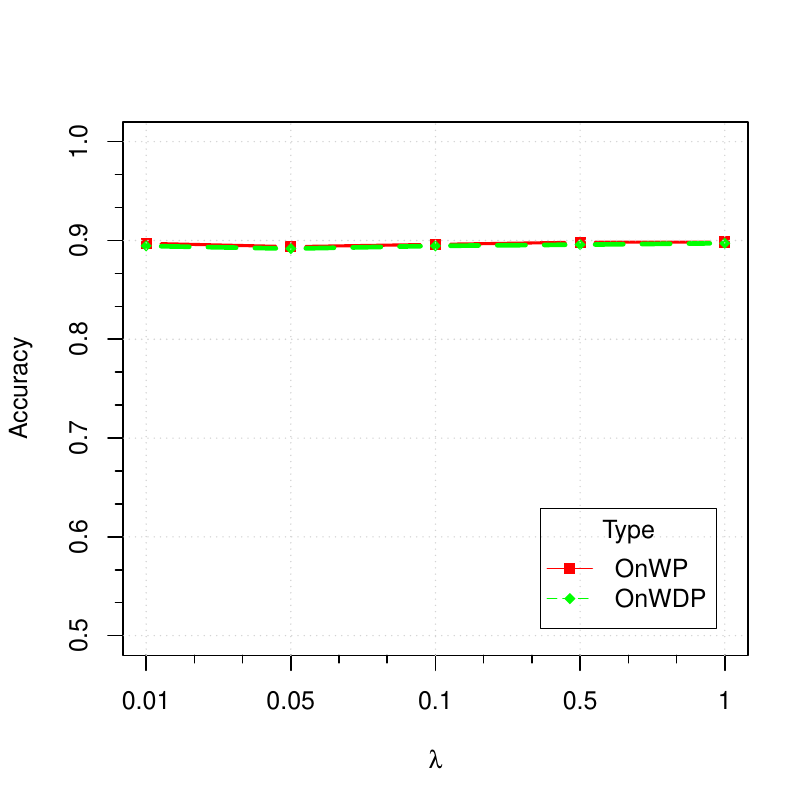}
        \caption{
}
        \label{fig:sub1}
    \end{subfigure}
    \hfill
    \begin{subfigure}[b]{0.48\textwidth}
        \centering
        \includegraphics[width=\textwidth]{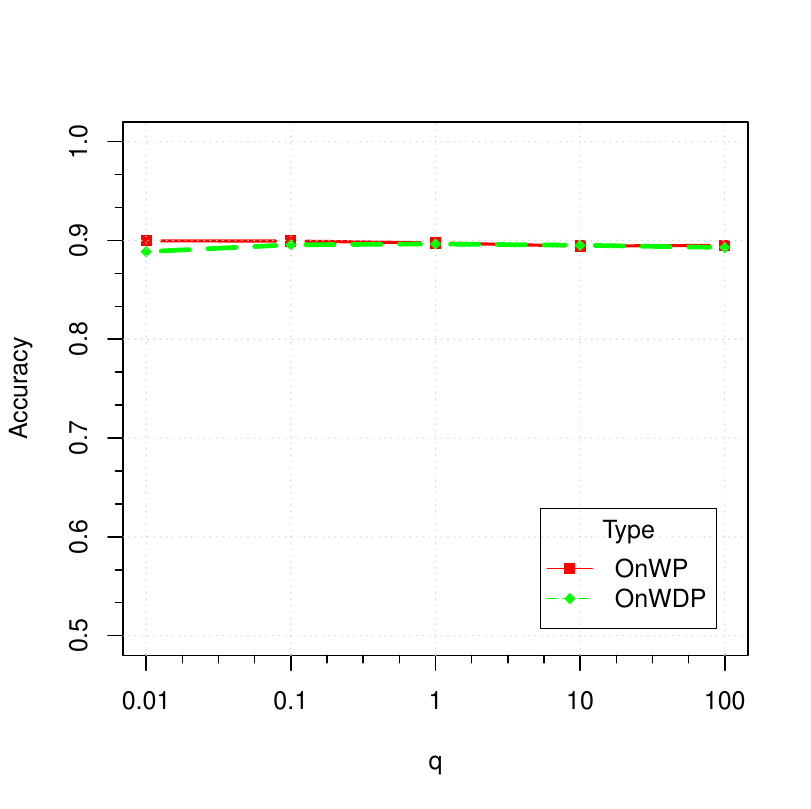}
        \caption{}
        \label{fig:sub2}
    \end{subfigure}
    \vfill
   \caption{Comparison of accuracy and time cost for the imbalanced dataset (two-class data, 4:1 ratio) based on variations in parameter $\lambda$ and $q$, with $M$=20, $p$=50, $\mu$=0.2, $b$=100, $\sigma$=1, $\epsilon=0.8$ and $\delta=10^{-5}$.}
    \label{fig7}
\end{figure}
\begin{figure}[htbp]
    \centering
    \begin{subfigure}[b]{0.48\textwidth}
        \centering
        \includegraphics[width=\textwidth]{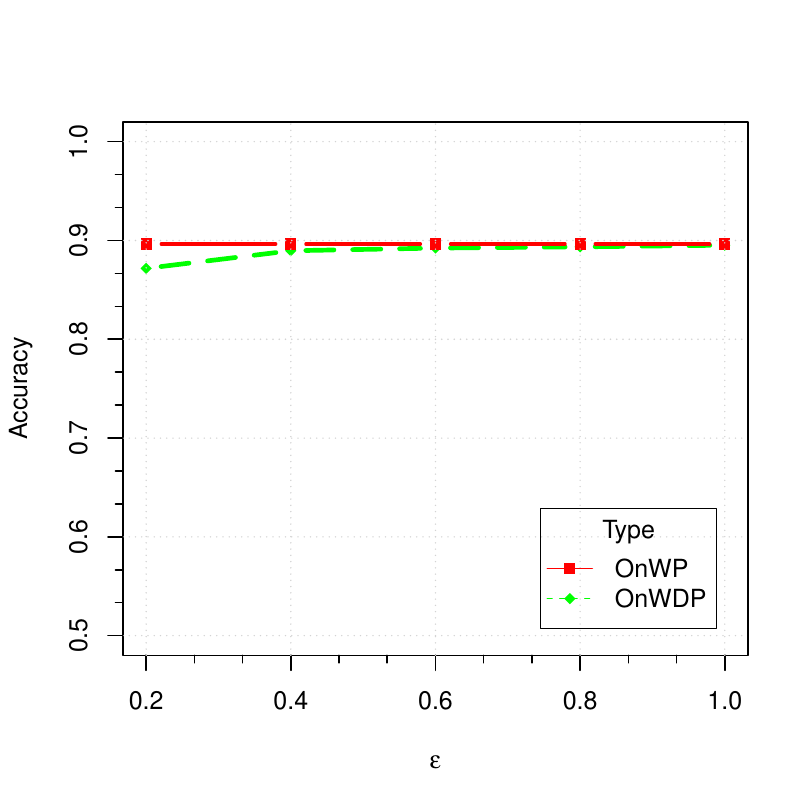}
        \caption{
}
        \label{fig:sub1}
    \end{subfigure}
    \hfill
    \begin{subfigure}[b]{0.48\textwidth}
        \centering
        \includegraphics[width=\textwidth]{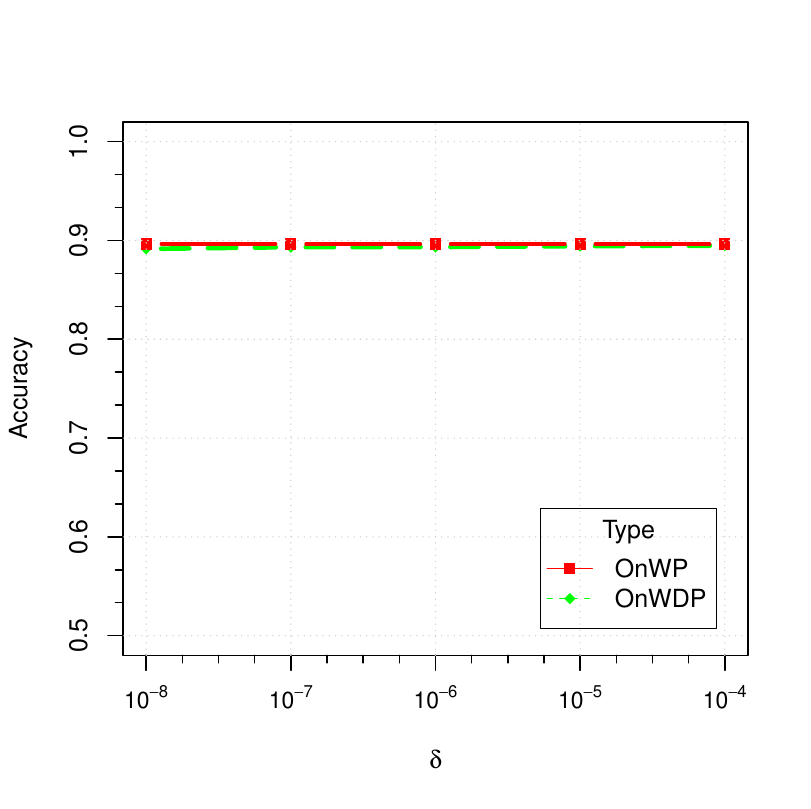}
        \caption{}
        \label{fig:sub2}
    \end{subfigure}
    \vfill
   \caption{Comparison of accuracy and time cost for the imbalanced dataset (two-class data, 4:1 ratio) based on variations in parameter $\epsilon$ and $\delta$, with $M$=20, $p$=50, $\mu$=0.2, $b$=100, $\sigma$=1 and $q=1$.}
    \label{fig8}
\end{figure}

Effect of parameter variations:
The effect of varying parameters such as $b$ (number of data batches), $M$ (number of sites), and $p$ (feature dimensions) demonstrates that OnWDP, OnWP and OffWP are more efficiently than OffNP. As the number of data batches ($b$) increases, OnWDP and OnWP continue to demonstrate its computational superiority, remaining nearly unaffected by increases in dataset size or complexity. In contrast, OffNP shows a noticeable slowdown in computation time as $b$ and $p$ increase, particularly as the number of sites grows.
When the feature dimensions $p$ vary,  OnWDP and OnWP again prove to be faster and more scalable, especially in large datasets. The privacy-preserving features of both methods seem to add minimal computational overhead. 
\\
Balanced vs. Imbalanced datasets:
The performance gap between the four methods become more significant in imbalanced datasets. In these cases, OffNP tends to struggle with lower accuracy, especially as the class imbalance increases (as seen in Figure \ref{fig6}). By contrast, OnWDP, OnWP and OffWP maintain significantly higher accuracy rates, even in highly imbalanced scenarios. This demonstrates the robustness of the proposed privacy-preserving methods in handling class imbalance, making them more suitable for real-world applications where data is often not evenly distributed.
In balanced datasets, the performance difference between the methods is smaller, with all achieving similar accuracy levels. Nevertheless, OnWDP and OnWP maintain its significantly faster computational speed, making it the ideal choice even when both classes are evenly represented.
﻿\\
We consider that $\mu$ and $\sigma$ vary between sites and are randomly generated from different Uniform distributions. From Table \ref{Tab3}, we can see that OnWDP and OnWP still achieve significantly faster computation times compared to both OffWP and OffNP, regardless of the scenario (whether balanced or imbalanced datasets).  In terms of accuracy, the differences between the methods are more significant in imbalanced datasets. For example, when $\mu$ and $\sigma$ are generated from $U(0, 0.3)$ and $U(0.1, 1)$, respectively, OnWDP and OnWP achieve accuracies around $72\%$, while OffWP still delivers higher accuracies. This indicates that OffWP offers a balance between computional speed and competitive accuracy. OffNP, however, shows lower accuracy, particularly in the imbalanced case.  Through Tables \ref{Tab1}-{Tab3}, we can see that the accuracy of the OnWDP method is slightly lower than that of the OnWP and OffWP methods. This is because stronger privacy protection measures are incorporated into the OnWDP method.
\\
Moreover, as shown in Figure \ref{fig7}, the online methods OnWDP and OnWP exhibit insensitivity to changes in the parameters $\lambda$ and $q$, highlighting the stability of these methods. 
In Figure \ref{fig8}(a), as the parameter $\epsilon$ varies, the accuracy for both OnWDP and OnWP remains nearly constant, indicating that their performance is largely unaffected by changes in $\epsilon$. When the value of $\epsilon$ approaches 0.2, a noticeable decrease for OnWDP in accuracy is observed. 
Similarly, in Figure \ref{fig8}(b), as the parameter $\delta$ changes, the accuracy remains stable for both methods, further demonstrating their robustness to variations in $\delta$.
\par
In summary, OnWDP and OnWP are the best options in terms of computation time, offering near-instant results even for large datasets, while maintaining accuracy levels similar to or better than OffNP. OnWDP shows a good balance between speed, accuracy and privacy preservation making it a strong alternative to the existing OffNP method, particularly in cases where class imbalance is an issue.

\begin{table}[h]
\scriptsize
\centering
\caption{Accuracy and computation time based on 100 simulation runs}
\label{Tab1}
\begin{tabular}{lcccccc}
\toprule
\multirow{2}{*}{Method} & \multicolumn{2}{c}{$M=10,b=100,p=50, \mu=0.2$} & \multicolumn{2}{c}{$M=10,b=1000,p=50, \mu=0.2$} & \multicolumn{2}{c}{$M=10,b=2000,p=50, \mu=0.2$}
\\
\cmidrule(r){2-3} \cmidrule(r){4-5}     \cmidrule(r){6-7}
&  Accuracy $(\%)$      &  Time (s) &  Accuracy $(\%)$
&  Time (s)      &  Accuracy $(\%)$  &  Time (s)
 \\
\midrule
& &  &   Balance
\\
OffNP     &92.1 (0.003)  &2.31  &92.1 (0.003) &27.01 &92.1 (0.003)&53.01
\\
OffWP    &92.1 (0.002) &2.14   &92.1 (0.003) &3.13 &92.2 (0.002) &9.57
\\
OnWP     &92.1 (0.003)  &0.10  &92.1 (0.003) &0.10&92.2 (0.003)&0.11 \\
OnWDP      &91.6 (0.002)  &0.10 &92.0 (0.003) &0.10&92.0 (0.003)&0.11\\
& &  &  Imbalance
\\
OffNP    &85.7 (0.003)  &2.79  &85.9 (0.003) &32.83 & 85.8 (0.004)  & 65.92
\\
OffWP     &90.2 (0.003)  &2.28   &90.3 (0.003) &3.24 &90.3 (0.003) &10.41 
\\
OnWP     &89.6 (0.003)  &0.11  &89.7 (0.003) &0.10&89.7 (0.003)&0.11 \\ 
OnWDP     &88.6 (0.005)  &0.11 &89.4 (0.003) &0.10&89.4 (0.003)&0.11\\ 
\bottomrule
\end{tabular}\\
`Balance' and `Imbalance' indicate that the ratio of the positive class to the negative class is 1:1 and 4:1, 
respectively. \\
\vspace{-3mm} The numbers in parentheses are the corresponding standard deviations. $q=1$, $\sigma=1$, $\epsilon=0.8$ and $\delta=10^{-5}$.
\end{table}

\begin{table}[h]
\scriptsize
\centering
\caption{Accuracy and computation time based on 100 simulation runs}
\label{Tab2}
\begin{tabular}{lcccccc}
\toprule
\multirow{2}{*}{Method} & \multicolumn{2}{c}{$M=50,b=100,p=10, \mu=0.2$} & \multicolumn{2}{c}{$M=50,b=100,p=20, \mu=0.2$} & \multicolumn{2}{c}{$M=50,b=100,p=100, \mu=0.2$}
\\
\cmidrule(r){2-3} \cmidrule(r){4-5}     \cmidrule(r){6-7}
&  Accuracy $(\%)$      &  Time (s) &  Accuracy $(\%)$
&  Time (s)      &  Accuracy $(\%)$  &  Time (s)
\\
\midrule
& &  &   Balance
\\
OffNP      &73.7 (0.004)  &2.87 &81.5 (0.004)  &4.58 &97.7 (0.002) &40.59
\\
OffWP     &73.7 (0.004)  &2.47   &81.5 (0.004) &4.12 &97.7 (0.002) & 26.73
\\
OnWP      &73.7 (0.004)  &0.21 &81.5 (0.004)  &0.28 &97.7 (0.001) &1.00 \\
OnWDP      &73.6 (0.004)  &0.21 &81.4 (0.004)  &0.28 &97.7 (0.001) &1.01 \\
& &  &  Imbalance
\\
OffNP      &52.6 (0.002)  &3.56 &66.2 (0.004)  &6.31 &95.6 (0.002) &42.78
\\
OffWP     &66.0 (0.003)  &2.67   &76.5 (0.004) &4.74 &97.2 (0.002) & 26.31
\\
OnWP     &61.0 (0.002)  &0.22 &74.2 (0.004)  &0.26 &97.1 (0.002) &0.97 \\
OnWDP     &60.9 (0.003)  &0.22 &74.2 (0.004)  &0.26 &97.0 (0.002) &0.98 \\
\bottomrule
\end{tabular}\\
`Balance' and `Imbalance' indicate that the ratio of the positive class to the negative class is 1:1 and 4:1, 
respectively. \\
\vspace{-3mm} The numbers in parentheses are the corresponding standard deviations. $q=1$, $\sigma=1$, $\epsilon=0.8$ and $\delta=10^{-5}$.
\end{table}

\begin{table}[h]
\scriptsize
\centering
\caption{Accuracy and computation time based on 100 simulation runs}
\label{Tab3}
\begin{tabular}{lcccccc}
\toprule
\multirow{2}{*}{Method} & \multicolumn{2}{c}{\makecell{$M=50,b=100,p=20,$\\$\mu\sim U(0, 0.3)$, $\sigma\sim U(0.1, 1)$}} & \multicolumn{2}{c}{\makecell{$M=50,b=100,p=20,$\\$\mu\sim U(0, 0.4)$, $\sigma\sim U(0.1, 1)$}} & \multicolumn{2}{c}{\makecell{$M=50,b=100,p=20,$\\$\mu\sim U(0.1, 0.4)$, $\sigma\sim U(0.1, 1)$}}
\\
\cmidrule(r){2-3} \cmidrule(r){4-5}     \cmidrule(r){6-7}
&  Accuracy $(\%)$      &  Time (s) &  Accuracy $(\%)$
&  Time (s)      &  Accuracy $(\%)$  &  Time (s)
\\
\midrule
& &  &   Balance
\\
OffNP     &81.2 (0.03)  &4.39 &88.3 (0.03)  &4.43 &93.5 (0.02) &4.98
\\
OffWP      &81.3 (0.03)  &4.33   &88.5 (0.03) &4.07 &93.6 (0.02) & 4.20
\\
OnWP     &81.2 (0.03)  &0.27 &88.3 (0.03)  &0.26 &93.5 (0.02) &0.27 \\
OnWDP      &81.1 (0.03)  &0.26 &88.2 (0.03)  &0.26 &93.4 (0.02) &0.27 \\
& &  &  Imbalance
\\
OffNP     &60.7 (0.02)  &5.68 &74.8 (0.03)  &5.28 &85.7 (0.02) &5.50
\\
OffWP      &78.4 (0.03)  &4.75   &85.8 (0.03) &4.63 &92.7 (0.02) & 4.25
\\
OnWP      &72.4 (0.03)  &0.27 &82.3 (0.03)  &0.26 &90.3 (0.02) &0.26 \\
OnWDP      &72.2 (0.04)  &0.27 &82.0 (0.03)  &0.26 &90.2 (0.03) &0.26 \\
\bottomrule
\end{tabular}\\
$U(\cdot, \cdot)$ denotes the uniform distribution. `Balance' and `Imbalance' indicate that the ratio of the positive class to the \\\vspace{-3mm}
negative class is 1:1 and 4:1, respectively. 
The numbers in parentheses are the corresponding standard deviations. $q=1$, \\\vspace{-3mm}$\epsilon=0.8$ and $\delta=10^{-5}$.

\end{table}

\newpage
\section{Real data study}
In this section,  we consider the Human Activity Recognition Trondheim (HARTH) dataset \citep{logacjov2021harth, bach2021machine}, which is available in the UCI Machine Learning Repository. The dataset can be accessed at the following link: \url{https://archive.ics.uci.edu/dataset/779/harth}.  
The HARTH dataset comprises data from 22 participants who each wore two Axivity AX3 accelerometers, both of which are 3-axis devices, for approximately 2 hours in a free-living setting. One sensor was positioned on the right front thigh, while the other was placed on the lower back. The data were collected at a sampling rate of 50 Hz. Additionally, video recordings from a chest-mounted camera were used to annotate the participants' activities on a frame-by-frame basis. The dataset contains the following annotated activities: walking, running, shuffling, stairs, standing, sitting, lying and cycling. 
\par
The dataset was created to develop and validate a new activity type recognition machine learning model (HARTH) that can accurately classify the physical behavior in daily life. This enhanced classification accuracy is particularly important for adults using walking aids. The goal is to provide a more precise description of physical activity levels in daily life, evaluate interventions, and understand the relationship between physical activity and health.
\\
The Human Activity Recognition data serve as an ideal benchmark for evaluating our proposed method due to its complexity and the collaborative, sustainable nature of the task. While it is relatively straightforward to detect walking activity when a smartphone is carried in a fixed location, real-life scenarios present more challenges. The device's orientation relative to the body and its position (whether in hands, bags, pockets, etc.) constantly change with movement, complicating detection.
\\
Moreover, no matter how comprehensive we believe our training data is, there will always be instances where specific users or devices prevent the model from generalizing effectively. Consequently, it is essential to continuously collect data from various participants, which introduces challenges related to data storage and privacy protection. Federated learning and online learning methods can avoid these issues by enabling decentralized data processing and continuous model updates.

\begin{figure}[h]  
  \centering  
  \includegraphics[width=0.8\textwidth]{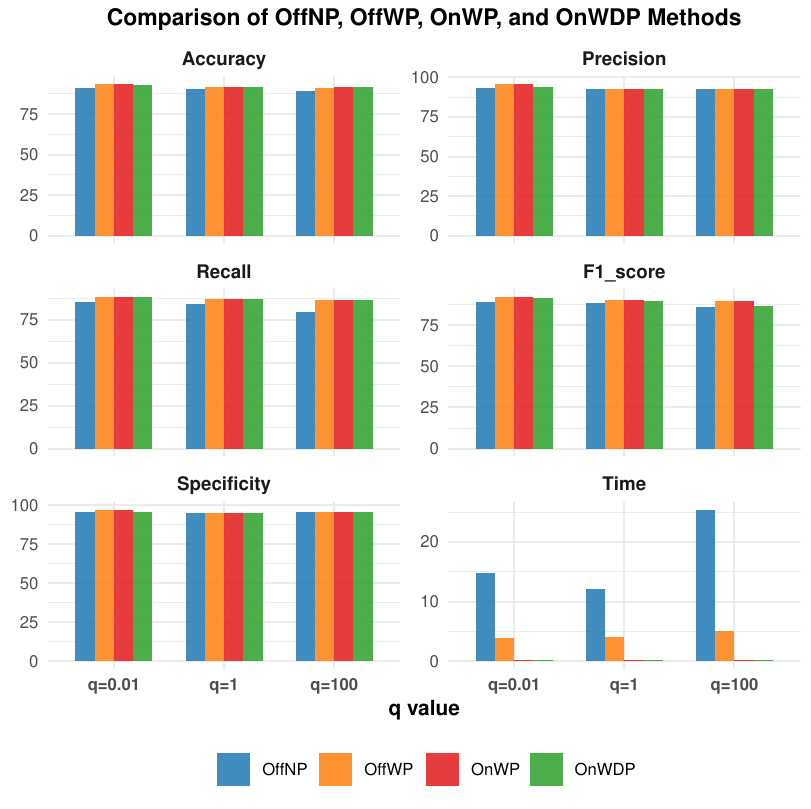}  
  \caption{A comparison of the prediction accuracy and computation time between the method OffNP and the methods OffWP, OnWP and OnWDP, based on different metrics and three different q values.}  
  \label{fig:example1}  
\end{figure}

We evaluate the proposed methods in a scenario where all annotated activities are classified into two broad categories: motion and rest. The motion category consists of activities involving movement, while the rest category includes states of inactivity. This scenario is designed to assess the methods' ability to distinguish between active and inactive states, which is essential for applications such as health monitoring and daily activity tracking. To provide a comprehensive assessment of the methods' performance, evaluation metrics such as accuracy, precision, recall, F1 score, specificity, and computation time are utilized. A random cross-validation strategy is employed to validate the generalizability and robustness of the models.
\\
From table \ref{Tab4} and figure \ref{fig:example1}, it is evident that the four methods—OffNP, OffWP, OnWP, and OnWDP—exhibit distinct trade-offs in terms of accuracy, computational efficiency, and privacy protection. Among these, OnWDP provides the strongest privacy protection, as it incorporates differential privacy mechanisms, making it the most suitable choice for applications where data confidentiality is paramount.  
\\
In terms of classification accuracy, OffWP and OnWP achieve the highest performance, particularly when $q=0.01$, indicating their effectiveness in distinguishing between `motion' and `rest' activities. OnWDP follows closely, demonstrating that strong privacy guarantees can still be maintained without significant accuracy degradation. Conversely, OffNP exhibits the lowest accuracy, particularly when $q=100$, suggesting that this method is more sensitive to variations in data distribution.  
\\
Considering computational efficiency, OnWP and OnWDP significantly outperform OffNP and OffWP, with execution times an order of magnitude lower. This efficiency makes them highly suitable for real-time applications. OffWP improves upon OffNP in terms of speed, but it remains computationally more expensive than the online methods. OffNP is the least efficient, especially for higher values of $q$, where processing time increases substantially.  
\\
Overall, the results indicate that OnWDP achieves the best balance between privacy preservation and computational efficiency while maintaining competitive accuracy. While OffWP yields the highest accuracy, it requires significantly more computational resources, making it less practical in time-sensitive scenarios. In contrast, OffNP lags behind in all three aspects, making it the least favorable option. These findings suggest that for applications requiring both strong privacy guarantees and efficient computation, OnWDP is the most robust and well-rounded choice.  

\begin{table}[h]
\footnotesize
\centering
\caption{\footnotesize Comparison of different methods for the activities `motion' and `rest' based on data split at random 100 times}
\label{Tab4}

\begin{tabular}{lcccccc}
\toprule Method &Accuracy $(\%)$& Precision $(\%)$& Recall $(\%)$& F1 score $(\%)$&Specificity $(\%)$&Time (s)\\
\hline OffNP$_{q=1}$  &90.8&  92.9& 84.4 & 88.4 &95.2& 12.16 \\
 OffNP$_{q=0.01}$  &91.3  &93.1 &85.4  &89.1 &95.5 &14.80 \\
 OffNP$_{q=100}$  &89.0  &92.9  & 79.7 & 85.8 &95.7 &25.43 \\
 OffWP$_{q=1}$  &91.9  &92.9  & 87.4 &90.0 &95.2 & 4.05 \\
 OffWP$_{q=0.01}$  &93.7  &95.9  & 88.4 &92.2 &97.3 &3.90  \\
 OffWP$_{q=100}$  &91.6  &92.9 & 86.7 &89.8  &95.6 &5.14\\
 OnWP$_{q=1}$  &91.9  &92.9  &87.3&89.9 &95.1&0.22  \\
 OnWP$_{q=0.01}$  &93.7  &95.8  &88.4  &92.0 &97.3 &0.20  \\
 OnWP$_{q=100}$  &91.6  &92.7  &86.6  &89.6  &95.6 &0.21 \\
 OnWDP$_{q=1}$  & 91.8 &92.9   &87.2   &89.8  &95.0  & 0.30  \\
 OnWDP$_{q=0.01}$  &92.9   &94.0   &88.5   &91.3  &96.0  & 0.28  \\
 OnWDP$_{q=100}$  &91.5   &92.5   & 86.5  &86.6   & 95.5 &0.29  \\
\bottomrule
\end{tabular}\\
The dataset is randomly divided into a training set and a testing set with a 4:1 ratio, $\epsilon=0.1$, $\delta=10^{-7}$.
\end{table}

\section{Conclusion}
In this article, we propose two online privacy-preserving methods based on the generalized Distance-Weighted Discriminant in federated learning environments: the online federated learning with differential privacy (OnWDP) and the online federated learning without differential privacy (OnWP). These methods are designed to efficiently handle streaming data from multiple decentralized clients while ensuring data privacy, computational efficiency, and model robustness. Unlike traditional batch-learning approaches, which require storing and retraining on entire datasets, our methods enable incremental learning, significantly reducing computational overhead and storage requirements, making them highly suitable for large-scale, real-time applications.
\\
A core contribution of our work is the introduction of renewable estimation procedures, allowing for efficient model updates without retaining past raw data, ensuring both efficiency and privacy. This approach significantly enhances the scalability and adaptability of federated learning, especially in settings where data arrives sequentially and centralized storage is not feasible. Theoretical analysis establishes that the proposed estimators achieve consistency, asymptotic normality, and Bayesian risk consistency, demonstrating their reliability and statistical robustness in real-world federated learning applications. Furthermore, our framework allows clients to collaborate in model training without sharing raw data, addressing major privacy concerns associated with decentralized learning.
Extensive empirical evaluations on both simulated and real-world datasets, including the Human Activity Recognition dataset, confirm that OnWP and OnWDP significantly improve classification performance, computational efficiency, and robustness. Among them, OnWP achieves the fastest model updates, making it ideal for real-time federated learning scenarios where rapid adaptation is crucial. Meanwhile, OnWDP effectively balances privacy and performance, incorporating differential privacy mechanisms to provide rigorous privacy protection while maintaining high classification accuracy. This ensures that OnWDP is particularly well-suited for privacy-sensitive applications.
Beyond privacy and computational efficiency, our proposed methods are highly scalable and can be seamlessly applied to diverse classification tasks, including those involving non-IID and imbalanced. The ability of OnWP and OnWDP to operate efficiently in dynamic, real-time environments makes them valuable tools for modern federated learning frameworks, particularly in scenarios where data privacy, adaptability, and computational constraints must be carefully balanced.

\FloatBarrier

\newpage

\bibliographystyle{chicago}
\bibliography{reference}

\section*{Appendix}
\par
\noindent
{\bf Proof of Theorem 1}:
\begin{proof}
Following \eqref{1-52} and \eqref{eq:3}, one has
\begin{eqnarray*}
H_D(\bm\theta)=\sum_{m=1}^M\sum_{i=1}^{n^{\scriptscriptstyle(m)}}\tilde V_q''(y_i\bar{\bm{x}}_i^T\bm{\theta})\bar{\bm{x}}_i\bar{\bm{x}}_i^T+ n^{\scriptscriptstyle(m)}\lambda \bm W\succcurlyeq 0,
\end{eqnarray*}
that is, $H_D(\bm\theta)$ is semi-positive definite. Therefore, $L_D(\bm\theta)$ is convex. As $\bar H_D(\bm\theta)- H_D(\bm\theta)\succcurlyeq 0$, the property of the convex function yields that
\begin{eqnarray*}
L_D(\bm\theta)\le L_D(\bm\theta')+(\bm\theta-\bm\theta')^T\nabla L_D(\bm\theta')+\frac12(\bm\theta-\bm\theta')^T\bar H_D(\bm\theta')(\bm\theta-\bm\theta'),
\end{eqnarray*}
for any $\bm\theta,\bm\theta'\in \mathbb{R}^{p+1}$. Then we have
\begin{eqnarray}\label{eq:5}
   L_D(\bm\theta)\le u(\bm\theta,\bm\theta'). 
\end{eqnarray}
By \eqref{eq:2}, it is easy to see that
\begin{eqnarray}\label{eq:6}
u(\bm\theta,\bm\theta)=L_D(\bm\theta)
\end{eqnarray}
As \eqref{D-5} is the minimization of $u(\bm\theta^{t+1},\bm\theta^t)$ given $\bm\theta^t$, by \eqref{eq:5} and \eqref{eq:6}, we have,
\begin{eqnarray*}
L_D(\bm\theta^{t+1})\le u(\bm\theta^{t+1},\bm\theta^t)\le u(\bm\theta^{t},\bm\theta^t)=L_D(\bm\theta^t).
\end{eqnarray*}
By induction,
\begin{eqnarray*}
L_D(\bm\theta^0)\ge L_D(\bm\theta^1)\ge L_D(\bm\theta^2)\ge...
\end{eqnarray*}
According to \eqref{1-4} and \eqref{eq:1}, the set $\{\bm\theta: L_D(\bm\theta)\le L_D(\bm\theta^0)\}$ is compact, then there exists a subsequence $\{\bm\theta^{r_j}\}$ in the set converging to a limit point $\bm\theta^*$. Then we have
\begin{eqnarray*}
u\left(\bm\theta^{r_{j+1}}, \bm\theta^{r_{j+1}}\right)=L_D\left(\bm\theta^{r_{j+1}}\right) \leq L_D\left(\bm\theta^{r_j+1}\right) \leq u\left(\bm\theta^{r_j+1}, \bm\theta^{r_j}\right) \leq u\left(\bm\theta, \bm\theta^{r_j}\right),
\end{eqnarray*}
for any $\bm\theta \in \mathbb{R}^{p+1}$. Let $j$ tends to infinity, one has
\begin{eqnarray*}
u(\hat{\bm\theta},\hat{\bm\theta})\le u(\bm\theta,\hat{\bm\theta}),    
\end{eqnarray*}
which implies
\begin{eqnarray*}
\nabla_{1,v}u(\hat{\bm\theta},\hat{\bm\theta})\ge \bm0.
\end{eqnarray*}
According to the definition of the direction derivative, 
\begin{eqnarray*}
\nabla_{1,\bm v}u(\hat{\bm\theta},\hat{\bm\theta})   
=\langle \nabla_{1}u(\hat{\bm\theta},\hat{\bm\theta}), \bm v  \rangle
=\langle \nabla L_D(\hat{\bm\theta}), \bm v  \rangle
=\nabla_{\bm v}L_D(\hat{\bm\theta}).
\end{eqnarray*}
Thus $\nabla_{\bm v}L_D(\hat{\bm\theta})\ge \bm0$, that is, $\hat{\bm\theta}$ is the stationary point of the loss function $L_D$. Since $L_D$ is convex, then we have that \eqref{D-5} converges to $\hat{\bm\theta}$.
As $\hat{\bm\theta}$ is an unbiased estimator of $\bm\theta^*$, then we have $\hat{\bm\theta}$ converges to $\bm\theta^*$ in probability as the sample size converges to infinity.

\end{proof}

{\bf Proof of Theorem 2}:
\begin{proof}
Setting the gradient of the objective function (\ref{Q-1}) to zero,  we have the following equation
\begin{equation}
\begin{aligned}
\bm\xi(D_b)=-N_b\nabla G(\bm\theta)-\rho\bm\theta.
\end{aligned}
\label{alproof-1}
\end{equation}
 It follows that
\begin{equation}
\begin{aligned}
\nabla\bm\xi(D_b)=-N_b\nabla^2 G(\bm\theta)-\rho\bm I_{p+1}.
\end{aligned}
\label{alproof-2}
\end{equation}
To prove that algorithm 2 satisfies $\epsilon$-differential privacy, it suffices to show that for all $\bm\alpha\in\mathbb R^{p+1}$,
$$
p(\tilde{\boldsymbol{\theta}}_b^p=\bm\alpha; D_b)\le e^{\epsilon}
p(\tilde{\boldsymbol{\theta}}_b^p=\bm\alpha; D_b').
$$
Note that 
\begin{equation}
\begin{aligned}
\frac{p(\tilde{\boldsymbol{\theta}}_b^p=\bm\alpha; D_b)}{p(\tilde{\boldsymbol{\theta}}_b^p=\bm\alpha; D_b')}
=\frac{p(\bm\xi|D_b)}{p(\bm\xi|D_b')}\cdot\frac{|\mathrm{det}(\nabla \bm\xi(D_b'))|}{|\mathrm{det}(\nabla \bm\xi(D_b))|},
\end{aligned}
\label{alproof-3}
\end{equation}
where $\mathrm{det}(\nabla \bm\xi(D_b'))$ and $\mathrm{det}(\nabla \bm\xi(D_b))$ denote 
Jacobian determinants under datasets $D_b$ and $D_b'$, respectively.
\\
Let 
$$-\bm P=\nabla \bm\xi(D_b)=
-\sum_{j=1}^b\sum_{m=1}^M
\bar H_{D_j}^m(\tilde{\bm\theta}_{j-1}^p)-\rho\bm I_{p+1}
$$
and
$$
-\bm Q=\tilde V_q^{\prime\prime}(y_{n_b}(1,\bm{x}_{n_b}^T)\tilde{\bm\theta}_{j-1}^p)(1,\bm{x}_{n_b}^T)^T(1,\bm{x}_{n_b}^T)-
\tilde V_q^{\prime\prime}(y_{n_b}'(1,\bm{x}_{n_b}'^T)\tilde{\bm\theta}_{j-1}^p)(1,\bm{x}_{n_b}'^T)^T(1,\bm{x}_{n_b}'^T).
$$
It implies that
$$
-(\bm P+\bm Q)=\nabla \bm\xi(D_b').
$$
Since 
$
\sum_{j=1}^b\sum_{m=1}^M\sum_{i=1}^{n_j^m}\tilde V_q^{\prime\prime}(y_{{\scriptscriptstyle D_{\! j}}i}(1,\bm{x}_{{\scriptscriptstyle D_{\! j}}i}^T)\tilde{\bm\theta}_{j-1}^p)
(1,\bm{x}_{{\scriptscriptstyle D_{\! j}}i}^T)^T(1,\bm{x}_{{\scriptscriptstyle D_{\! j}}i}^T)
$ is positive semi-definite matrix, without loss of generality, assuming \( \lambda_j\)'s are equal, we have that 
 the smallest eigenvalue of $\bm P$ is $N_b\lambda+\rho$. It follows that
\begin{equation}
\begin{aligned}
|\gamma_k(\bm P^{-1}\bm Q)|\le \frac{|\gamma_k(\bm Q)|}{N_b\lambda+\rho},~k=1,2.
\end{aligned}
\label{alproof-4}
\end{equation}
Note that
\begin{equation}
\begin{aligned}
&\mathrm{tr}(\bm Q) = \mathrm{tr}[-\tilde V_q^{\prime\prime}(y_{n_b}(1,\bm{x}_{n_b}^T)\tilde{\bm\theta}_{j-1}^p)(1,\bm{x}_{n_b}^T)^T(1,\bm{x}_{n_b}^T)+\tilde V_q^{\prime\prime}(y_{n_b}'(1,\bm{x}_{n_b}'^T)\tilde{\bm\theta}_{j-1}^p)(1,\bm{x}_{n_b}'^T)^T(1,\bm{x}_{n_b}'^T)].
\end{aligned}
\label{alproof-5}
\end{equation}
Using (\ref{alproof-5}), $\tilde V_q^{\prime\prime}(\cdot)\le\frac{(q+1)^2}{q}$, $\|(1,\bm{x}_{n_b}^T)^T\|\le C_2$ and $\|(1,\bm{x}_{n_b}'^T)^T\|\le C_2$,
we have the following inequality
\begin{equation}
\begin{aligned}
|\gamma_1(\bm Q)|+|\gamma_2(\bm Q)|
&\le \tilde V_q^{\prime\prime}(y_{n_b}(1,\bm{x}_{n_b}^T)\tilde{\bm\theta}_{j-1}^p)\|(1,\bm{x}_{n_b}^T)^T\|^2+\tilde V_q^{\prime\prime}(y_{n_b}'(1,\bm{x}_{n_b}'^T)\tilde{\bm\theta}_{j-1}^p)\|(1,\bm{x}_{n_b}'^T)^T\|^2\\
&\le \frac{2(q+1)^2C_2^2}{q}.
\end{aligned}
\label{alproof-6}
\end{equation}
Combining (\ref{alproof-4}) and (\ref{alproof-6}) yields 
\begin{equation}
\begin{aligned}
|\gamma_1(\bm P^{-1}\bm Q)|+|\gamma_2(\bm P^{-1}\bm Q)|\le \frac{2(q+1)^2C_2^2}{(N_b\lambda+\rho)q}.
\end{aligned}
\label{alproof-7}
\end{equation}
Further, by (\ref{lemma-1}) and (\ref{alproof-7}), we have
\begin{equation}
\begin{aligned}
\frac{|\operatorname{det}(\bm {P}+\bm {Q})|}{|\operatorname{det}(\bm P)|}
\le 1+ \frac{2(q+1)^2C_2^2}{(N_b\lambda+\rho)q}+ \frac{(q+1)^4C_2^4}{[(N_b\lambda+\rho)q]^2}
= \left[1+ \frac{(q+1)^2C_2^2}{(N_b\lambda+\rho)q}\right]^2.
\end{aligned}
\label{alproof-8}
\end{equation}
Since
$\bm\xi \in \mathbb{R}^{p+1}$ is a random vector with Laplace density function 
$p(\bm\xi)=\frac{1}{(2\eta)^{p+1}} e^{-\|\bm\xi\|_1 /\eta}$, we have
\begin{equation}
\begin{aligned}
\frac{p(\bm\xi|D_b)}{p(\bm\xi|D_b')}=\frac{e^{-\|\bm\xi(D_b)\|_1 /\eta}}{e^{-\|\bm\xi(D_b')\|_1 /\eta}}
= \mathrm{exp}\left\{\frac{1}{\eta}(\|\bm\xi(D_b')\|_1 -\|\bm\xi(D_b)\|_1 )\right\}.
\end{aligned}
\label{alproof-9}
\end{equation}
Following the norm triangle inequality, we have
\begin{equation}
\begin{aligned}
&\|\bm\xi(D_b')\|_1 -\|\bm\xi(D_b)\|_1 \\
&\le \|\bm\xi(D_b')-\bm\xi(D_b)\|_1\\
&=\|N_b\nabla G(D_b)- N_b\nabla G(D_b')\|_1\\
&=\|\nabla L_{D_b}(\tilde{\boldsymbol{\theta}}_{b-1}^p)+\sum_{j=1}^b \bar{H}_{D_j}(\tilde{\boldsymbol{\theta}}_{j-1}^p)(\boldsymbol{\theta}-\tilde{\boldsymbol{\theta}}_{b-1}^p)
-\nabla L_{D_b'}(\tilde{\boldsymbol{\theta}}_{b-1}^p)-\sum_{j=1}^b \bar{H}_{D_j'}(\tilde{\boldsymbol{\theta}}_{j-1}^p)(\boldsymbol{\theta}-\tilde{\boldsymbol{\theta}}_{b-1}^p)\|_1\\
&=\|[y_{n_b}V_q^{\prime}(y_{n_b}\bar{\bm{x}}_{n_b}^T\tilde{\bm\theta}_{b-1}^p)\bar{\bm{x}}_{n_b}-
y_{n_b}'V_q^{\prime}(y_{n_b}'\bar{\bm{x}}_{n_b}'^T\tilde{\bm\theta}_{b-1}^p)\bar{\bm{x}}_{n_b}']\\
&~~~~+[\tilde V_q^{\prime\prime}(y_{n_b}\bar{\bm{x}}_{n_b}^T\tilde{\bm\theta}_{b-1}^p)\bar{\bm{x}}_{n_b}\bar{\bm{x}}_{n_b}^T-\tilde V_q^{\prime\prime}(y_{n_b}'\bar{\bm{x}}_{n_b}'^T\tilde{\bm\theta}_{b-1}^p)\bar{\bm{x}}_{n_b}'\bar{\bm{x}}_{n_b}'^T](\boldsymbol{\theta}-\tilde{\boldsymbol{\theta}}_{b-1}^p)\|_1\\
&\le
\|y_{n_b}V_q^{\prime}(y_{n_b}\bar{\bm{x}}_{n_b}^T\tilde{\bm\theta}_{b-1}^p)\bar{\bm{x}}_{n_b}\|_1+
\|y_{n_b}'V_q^{\prime}(y_{n_b}'\bar{\bm{x}}_{n_b}'^T\tilde{\bm\theta}_{b-1}^p)\bar{\bm{x}}_{n_b}'\|_1\\
&~~~~+\|\tilde V_q^{\prime\prime}(y_{n_b}\bar{\bm{x}}_{n_b}^T\tilde{\bm\theta}_{b-1}^p)\bar{\bm{x}}_{n_b}\bar{\bm{x}}_{n_b}^T(\boldsymbol{\theta}-\tilde{\boldsymbol{\theta}}_{b-1}^p)\|_1+\|\tilde V_q^{\prime\prime}(y_{n_b}'\bar{\bm{x}}_{n_b}'^T\tilde{\bm\theta}_{b-1}^p)\bar{\bm{x}}_{n_b}'\bar{\bm{x}}_{n_b}'^T(\boldsymbol{\theta}-\tilde{\boldsymbol{\theta}}_{b-1}^p)\|_1.
\end{aligned}
\label{alproof-10}
\end{equation}
Recall that
$|V_q^{\prime}(\cdot)|\le 1$, $|y_{n_b}|\le 1$, $|\tilde V_q^{\prime\prime}(\cdot)|\le \frac{(q+1)^2}{q}$, $|y_{n_b}|\le 1$, $|y_{n_b}'|\le 1$, $\|\bar{\bm{x}}_{n_b}\|_1\le C_1$, $\|\bar{\bm{x}}_{n_b}'\|_1\le C_1$,
$\|\bar{\bm{x}}_{n_b}\|\le C_2$, $\|\bar{\bm{x}}_{n_b}'\|\le C_2$ and
$\|\bar{\bm{x}}_{n_b}\bar{\bm{x}}_{n_b}^T(\boldsymbol{\theta}-\tilde{\boldsymbol{\theta}}_{b-1}^p)\|_1
\le \|\bar{\bm{x}}_{n_b}\|_1\|\bar{\bm{x}}_{n_b}\|\|\boldsymbol{\theta}-\tilde{\boldsymbol{\theta}}_{b-1}^p\|$, $\|\bar{\bm{x}}_{n_b}'\bar{\bm{x}}_{n_b}'^T(\boldsymbol{\theta}-\tilde{\boldsymbol{\theta}}_{b-1}^p)\|_1\le \|\bar{\bm{x}}_{n_b}'\|_1\|\bar{\bm{x}}_{n_b}'\|\|\boldsymbol{\theta}-\tilde{\boldsymbol{\theta}}_{b-1}^p\|$,
$\|\boldsymbol{\theta}-\tilde{\boldsymbol{\theta}}_{b-1}^p\|=C_{p-1}/\sqrt{N_{b-1}}$, $C_{p-1}=O_p(1)$.
\\
Moreover, by (\ref{alproof-10}), it follows that
\begin{equation}\nonumber
\begin{aligned}
&\|\bm\xi(D_b')\|_1 -\|\bm\xi(D_b)\|_1 \le 2C_1+\frac{2(q+1)^2C_1C_2C_{p-1}}{q\sqrt{N_{b-1}}}.
\end{aligned}
\label{alproof-11}
\end{equation}
Thus, let 
\begin{equation}
\begin{aligned}
T_1&=2C_1+\frac{2(q+1)^2C_1C_2C_{p-1}}{q\sqrt{N_{b-1}}}\\
T_2&= 2\mathrm{ln}\left(1+\frac{(q+1)^2 C_2^2}{\left(N_b \lambda+\rho\right) q}\right),
\end{aligned}
\label{alproof-11-1}
\end{equation}
we have
\begin{equation}
\begin{aligned}
\frac{p(\tilde{\boldsymbol{\theta}}_b^p=\bm\alpha; D_b)}{p(\tilde{\boldsymbol{\theta}}_b^p=\bm\alpha; D_b')}
&=\frac{p(\bm\xi|D_b)}{p(\bm\xi|D_b')}\cdot\frac{|\mathrm{det}(\nabla \bm\xi(D_b'))|}{|\mathrm{det}(\nabla \bm\xi(D_b))|}\\
&\le
\mathrm{exp}\left\{\frac{1}{\eta}\left(2C_1+\frac{2(q+1)^2C_1C_2C_{p-1}}{q\sqrt{N_{b-1}}}\right)+ 2\mathrm{ln}\left(1+\frac{(q+1)^2 C_2^2}{\left(N_b \lambda+\rho\right) q}\right)\right\}\\
&=\mathrm{exp}\left\{\frac{1}{\eta}T_1+ T_2\right\}.
\end{aligned}
\label{alproof-12}
\end{equation}
If $\eta=(\epsilon-T_2)^{-1}T_1$, then
$
\frac{p(\tilde{\boldsymbol{\theta}}_b^p=\bm\alpha; D_b)}{p(\tilde{\boldsymbol{\theta}}_b^p=\bm\alpha; D_b')}
\le e^{\epsilon}.
$
Algorithm 2 is $\epsilon$-differential privacy.
\par
Next, to prove that algorithm \( \mathcal{A} \) satisfies \((\epsilon, \delta)\)-differential privacy, we let 
$\Delta\bm\xi=\bm\xi(D_b')-\bm\xi(D_b)$. It implies that 
\begin{equation}
\begin{aligned}
\|\bm\xi(D_b')\|^2-\|\bm\xi(D_b)\|^2=\|\Delta\bm\xi+\bm\xi(D_b)\|^2-\|\bm\xi(D_b)\|^2=2\bm\xi(D_b)^T\Delta\bm\xi+\|\Delta\bm\xi\|^2.
\end{aligned}
\label{alproof-13}
\end{equation}
Therefore, similar to (\ref{alproof-10}), we obtain
\begin{equation}
\begin{aligned}
\|\Delta\bm\xi\|&=\|\bm\xi(D_b')-\bm\xi(D_b)\|\\
&\le
\|y_{n_b}V_q^{\prime}(y_{n_b}\bar{\bm{x}}_{n_b}^T\tilde{\bm\theta}_{b-1}^p)\bar{\bm{x}}_{n_b}\|+
\|y_{n_b}'V_q^{\prime}(y_{n_b}'\bar{\bm{x}}_{n_b}'^T\tilde{\bm\theta}_{b-1}^p)\bar{\bm{x}}_{n_b}'\|\\
&~~~~+\|\tilde V_q^{\prime\prime}(y_{n_b}\bar{\bm{x}}_{n_b}^T\tilde{\bm\theta}_{b-1}^p)\bar{\bm{x}}_{n_b}\bar{\bm{x}}_{n_b}^T(\boldsymbol{\theta}-\tilde{\boldsymbol{\theta}}_{b-1}^p)\|+\|\tilde V_q^{\prime\prime}(y_{n_b}'\bar{\bm{x}}_{n_b}'^T\tilde{\bm\theta}_{b-1}^p)\bar{\bm{x}}_{n_b}'\bar{\bm{x}}_{n_b}'^T(\boldsymbol{\theta}-\tilde{\boldsymbol{\theta}}_{b-1}^p)\|\\
&\le 2C_2+\frac{2(q+1)^2C_2^2C_{p-1}}{q\sqrt{N_{b-1}}}\triangleq\Delta_1.
\end{aligned}
\label{alproof-14}
\end{equation}
Furthermore, we partition $\mathbb{R}^{p+1}$ into two subsets $S_1$ and $S_2$ such as $\mathbb{R}^{p+1}=S_1\cup S_2$, where 
$S_1=\{\bm\xi\in\mathbb{R}^{p+1}|\bm\xi^T\Delta\bm\xi\le\Delta_1\tau t\}$ and 
$S_2=\{\bm\xi\in\mathbb{R}^{p+1}|\bm\xi^T\Delta\bm\xi>\Delta_1\tau t\}$. 
\\
Using the fact that for
any random variable $X$ following $\mathcal{N}\left(0, \tau^2\right)$ satisfies the upper deviation inequality
$$
\mathbb{P}[X \geq t] \leq e^{-\frac{t^2}{2 \tau^2}} \quad \text { for all } t \geq 0
$$
and the fact that
if the vector $\bm\xi\sim \mathcal{N}\left(\bm 0, \tau^2\bm I_{p+1}\right),$ then
$$
\bm\xi^T\Delta\bm\xi\sim \mathcal{N}\left(\bm 0, \|\Delta\bm\xi\|^2\tau^2\right),
$$
we have
$
\operatorname{Pr}(\bm\xi^T\Delta\bm\xi\ge\|\Delta\bm\xi\|\tau t)\le e^{-\frac{t^2}{2}},
$
In addition, let $\delta=e^{-\frac{t^2}{2}}$, then $t=\sqrt{2\mathrm{ln}\frac{1}{\delta}}$. This implies that
\begin{equation}
\begin{aligned}
\operatorname{Pr}(\bm\xi^T\Delta\bm\xi>\Delta_1\tau t)\le \delta.
\end{aligned}
\label{alproof-15}
\end{equation}
Next, if the vector $\bm\xi\sim \mathcal{N}\left(\bm 0, \tau^2\bm I_{p+1}\right)$, then 
\begin{equation}
\begin{aligned}
\frac{p(\bm\xi|D_b)}{p(\bm\xi|D_b')}=\frac{e^{-\|\bm\xi(D_b)\|^2 /2\tau^2}}{e^{-\|\bm\xi(D_b')\|^2/2\tau^2}}
= \mathrm{exp}\left\{\frac{1}{2\tau^2}(\|\bm\xi(D_b')\|^2 -\|\bm\xi(D_b)\|^2 )\right\}.
\end{aligned}
\label{alproof-16}
\end{equation}
Further, we consider $\bm\xi\in S_1$. Using (\ref{alproof-13}), (\ref{alproof-14}) and (\ref{alproof-16}) yielding
\begin{equation}
\begin{aligned}
\frac{p(\bm\xi\in S_1|D_b)}{p(\bm\xi\in S_1|D_b')}\le \mathrm{exp}\left\{\frac{1}{2\tau^2}\Big(\Delta_1\tau\sqrt{8\mathrm{ln}\frac{1}{\delta}}+\Delta_1^2\Big)\right\}.
\end{aligned}
\label{alproof-17}
\end{equation}
Combining (\ref{alproof-3}), (\ref{alproof-8}) and (\ref{alproof-17}), we have
\begin{equation}
\begin{aligned}
\frac{p(\tilde{\boldsymbol{\theta}}_b^p=\bm\alpha|\bm\xi\in S_1; D_b)}{p(\tilde{\boldsymbol{\theta}}_b^p=\bm\alpha|\bm\xi\in S_1; D_b')}
\le \exp \left\{\frac{1}{2\tau^2}\Big(\Delta_1\tau\sqrt{8\mathrm{ln}\frac{1}{\delta}}+\Delta_1^2\Big)+2 \ln \left(1+\frac{(q+1)^2 C_2^2}{\left(N_b \lambda+\rho\right) q}\right)\right\}.
\end{aligned}
\label{alproof-18}
\end{equation}
Let
$$
\frac{1}{2\tau^2}\Big(\Delta_1\tau\sqrt{8\mathrm{ln}\frac{1}{\delta}}+\Delta_1^2\Big)=\frac{\epsilon}{2}
$$
and
$$
2 \ln \left(1+\frac{(q+1)^2 C_2^2}{\left(N_b \lambda+\rho\right) q}\right)=\frac{\epsilon}{2}.
$$
By solving the above two equations, we get
$$
\tau=\Delta_1\left(\sqrt{2\mathrm{ln}\frac{1}{\delta}}+\sqrt{2\mathrm{ln}\frac{1}{\delta}+\epsilon}\right)\epsilon^{-1}
$$
and
$$
\epsilon=4 \ln \left(1+\frac{(q+1)^2 C_2^2}{\left(N_b \lambda+\rho\right) q}\right).
$$
This implies that the inequality (\ref{alproof-18}) can be expressed as 
\begin{equation}
\begin{aligned}
\frac{p(\tilde{\boldsymbol{\theta}}_b^p=\bm\alpha|\bm\xi\in S_1; D_b)}{p(\tilde{\boldsymbol{\theta}}_b^p=\bm\alpha|\bm\xi\in S_1; D_b')}
\le e^{\epsilon}.
\end{aligned}
\label{alproof-19}
\end{equation}
Thus, by (\ref{alproof-15}) and (\ref{alproof-19}), we have
\begin{equation}\nonumber
\begin{aligned}
p(\tilde{\boldsymbol{\theta}}_b^p=\bm\alpha; D_b)=&
p(\tilde{\boldsymbol{\theta}}_b^p=\bm\alpha|\bm\xi\in S_1; D_b)Pr(\bm\xi\in S_1)\\
&+p(\tilde{\boldsymbol{\theta}}_b^p=\bm\alpha|\bm\xi\in S_2; D_b)Pr(\bm\xi\in S_2)\\
&\le e^{\epsilon}p(\tilde{\boldsymbol{\theta}}_b^p=\bm\alpha|\bm\xi\in S_1; D_b')Pr(\bm\xi\in S_1)+\delta\\
&= e^{\epsilon}p(\tilde{\boldsymbol{\theta}}_b^p=\bm\alpha; D_b')+\delta.
\end{aligned}
\label{alproof-20}
\end{equation}
This completes the proof of theorem 2.

\end{proof}
\par
\noindent
{\bf Proof of Proposition 1}:
\begin{proof}
For the sake of convenience in proving, introduce a function
\begin{eqnarray}
g(\bm\theta)=\frac{1}{N_b}\sum_{j=1}^{b}J_j(D_j;\tilde{\bm\theta}_j^p)(\bm\theta-\tilde{\bm\theta}_{b-1}^p)+ \frac{1}{N_b}\nabla L_{D_b}(\bm\theta_b^0)+\frac{1}{N_b}\rho\bm\theta+\frac{1}{N_b}\bm\xi.
\label{proof2-1}
\end{eqnarray}
Setting $\bm\theta_b^0=\tilde{\bm\theta}_{b-1}^p$, the estimator $\tilde{\bm\theta}_b^p$ satisfies
\begin{eqnarray}
g(\tilde{\bm\theta}_b^p)=\frac{1}{N_b}\sum_{j=1}^{b}J_j(D_j;\tilde{\bm\theta}_j^p)(\tilde{\bm\theta}_b^p-\tilde{\bm\theta}_{b-1}^p)+ \frac{1}{N_b}\nabla L_{D_b}(\tilde{\bm\theta}_{b-1}^p)+\frac{1}{N_b}\rho\bm\tilde{\bm\theta}_b^p+\frac{1}{N_b}\bm\xi=\bm 0.
\label{proof2-2}
\end{eqnarray}
If $\{\tilde{\bm\theta}_j\}_{j=1}^{b-1}$ are consistent, we have
\begin{eqnarray}
g(\bm\theta_0)=\frac{1}{N_b}\sum_{j=1}^{b}J_j(D_j;\tilde{\bm\theta}_j^p)(\bm\theta_0-\tilde{\bm\theta}_{b-1}^p)+ \frac{1}{N_b}\nabla L_{D_b}(\tilde{\bm\theta}_{b-1}^p)+\frac{1}{N_b}\rho\bm\theta_0+\frac{1}{N_b}\bm\xi=o_p(1).
\label{proof2-3}
\end{eqnarray}
Thus, combining equations (\ref{proof2-2}) and (\ref{proof2-3}) results in
\begin{eqnarray}
g(\tilde{\bm\theta}_b^p)-g(\bm\theta_0)=\frac{1}{N_b}\sum_{j=1}^{b}J_j(D_j;\tilde{\bm\theta}_j^p)(\tilde{\bm\theta}_b^p-\bm\theta_0)+\frac{1}{N_b}\rho(\tilde{\bm\theta}_b^p-\bm\theta_0)=o_p(1).
\label{proof2-4}
\end{eqnarray}
By $\frac{1}{N_b}\sum_{j=1}^{b}J_j(D_j;\tilde{\bm\theta}_j^p)$ is positive definite, we have that $\tilde{\bm\theta}_{b}^p\xrightarrow{\mathrm{P}}\bm{\theta}_0$, as $N_b\rightarrow\infty$.

\end{proof}
\par
\noindent
{\bf Proof of Proposition 2}:
\begin{proof}
Using the equations (\ref{proof2-1}), (\ref{proof2-2}) and (\ref{proof2-4}) yields
\begin{equation}
\begin{aligned}
g(\bm\theta_0)&=\frac{1}{N_b}\sum_{j=1}^{b}J_j(D_j;\tilde{\bm\theta}_j^p)(\bm\theta_0-\tilde{\bm\theta}_{b-1}^p)+ \frac{1}{N_b}\nabla L_{D_b}(\tilde{\bm\theta}_{b-1}^p)+\frac{1}{N_b}\rho\bm\theta_0+\frac{1}{N_b}\bm\xi
\\
&=-\frac{1}{N_b}\sum_{j=1}^{b}J_j(D_j;\tilde{\bm\theta}_j^p)(\tilde{\bm\theta}_{b}^p-\bm\theta_0)-\frac{1}{N_b}\rho(\tilde{\bm\theta}_{b}^p-\bm\theta_0)=o_p(1).
\end{aligned}
\label{th3proof1}
\end{equation}
It implies that
\begin{equation}
\begin{aligned}
&\frac{1}{N_b}\sum_{j=1}^{b}J_j(D_j;\tilde{\bm\theta}_j^p)(\tilde{\bm\theta}_{b}^p-\bm\theta_0)+\frac{1}{N_b}\sum_{j=1}^{b}J_j(D_j;\tilde{\bm\theta}_j^p)(\bm\theta_0-\tilde{\bm\theta}_{b-1}^p)+ \frac{1}{N_b}\nabla L_{D_b}(\tilde{\bm\theta}_{b-1}^p)\\
&+\frac{1}{N_b}\rho(\tilde{\bm\theta}_{b}^p-\bm\theta_0)+\frac{1}{N_b}\rho\bm\theta_0+\frac{1}{N_b}\bm\xi
=\bm 0.
\end{aligned}
\label{th3proof2}
\end{equation}
Taking the first-order Taylor series expansion of $\nabla L_{D_j}(\bm\theta)$ around $\tilde{\boldsymbol{\theta}}_{j}^p$, we have 
\begin{equation}
\begin{aligned}
\frac{1}{N_b}\sum_{j=1}^b\nabla L_{D_j}(\bm\theta_0)=&\frac{1}{N_b}\sum_{j=1}^b\nabla L_{D_j}(\tilde{\bm\theta}_{j}^p)+\frac{1}{N_b}\sum_{j=1}^{b}J_j(D_j;\tilde{\bm\theta}_j^p)(\bm\theta_0-\tilde{\bm\theta}_{j}^p)\\
&+O_p\Big(\sum_{j=1}^b\frac{n_j}{N_b}\|\boldsymbol{\theta}_0-\tilde{\boldsymbol{\theta}}_j^p\|^2\Big).
\end{aligned}
\label{th3proof3}
\end{equation}
Since $\tilde{\bm\theta}_j^p$ is a consistent estimator of $\bm\theta_0$, we know that
$\frac{1}{N_b}\sum_{j=1}^b\nabla L_{D_j}(\tilde{\bm\theta}_{j}^p)=o_p(1)$ and $\frac{1}{N_b}\nabla L_{D_b}(\tilde{\bm\theta}_{b-1}^p)=o_p(1)$.
It follows that
\begin{equation}
\begin{aligned}
\frac{1}{N_b}\sum_{j=1}^b\nabla L_{D_j}(\bm\theta_0)=
\frac{1}{N_b}\sum_{j=1}^{b}J_j(D_j;\tilde{\bm\theta}_j^p)(\bm\theta_0-\tilde{\bm\theta}_{j}^p)+O_p\Big(\sum_{j=1}^b\frac{n_j}{N_b}\|\boldsymbol{\theta}_0-\tilde{\boldsymbol{\theta}}_j^p\|^2\Big).
\end{aligned}
\label{th3proof4}
\end{equation}
From (\ref{th3proof2}) and (\ref{th3proof4}), it is easy to see that 
\begin{equation}
\begin{aligned}
&\frac{1}{N_b}\sum_{j=1}^b\nabla L_{D_j}(\bm\theta_0)+
\frac{1}{N_b}\sum_{j=1}^{b}J_j(D_j;\tilde{\bm\theta}_j^p)(\tilde{\bm\theta}_{b}^p-\bm\theta_0)+\frac{1}{N_b}\sum_{j=1}^{b}J_j(D_j;\tilde{\bm\theta}_j^p)(\tilde{\bm\theta}_{j}^p-\tilde{\bm\theta}_{b-1}^p)\\
&+\frac{1}{N_b}\rho\bm\theta_0+\frac{1}{N_b}\bm\xi+
O_p\Big(\sum_{j=1}^b\frac{n_j}{N_b}\|\boldsymbol{\theta}_0-\tilde{\boldsymbol{\theta}}_j^p\|^2\Big)=0
\end{aligned}
\label{th3proof5}
\end{equation}
By the weak law of large numbers, we know that 
$\frac{1}{N_b}\sum_{j=1}^{b}J_j(D_j;\bm\theta_0)\xrightarrow{\mathrm{P}} \bm\Sigma_0^{-1}$, as $N_b\rightarrow\infty$.
Furthermore, by $\tilde{\bm\theta}_j^p$ is a consistent estimator, $\bm\theta_0$ =O(1) and the continuous mapping theorem, we have 
\begin{equation}
\begin{aligned}
\frac{1}{\sqrt{N_b}}\sum_{j=1}^b\nabla L_{D_j}(\bm\theta_0)+
\frac{1}{\sqrt{N_b}}\sum_{j=1}^{b}J_j(D_j;\bm\theta_0)(\tilde{\bm\theta}_{b}^p-\bm\theta_0)+o_p(1)=0
\end{aligned}
\label{th3proof6}
\end{equation}
Equation (\ref{th3proof6}) is rewritten as
\begin{equation}
\begin{aligned}
\sqrt{N_b}(\bm\theta_0-\tilde{\bm\theta}_{b}^p)=\Big[\frac{1}{N_b}\sum_{j=1}^{b}J_j(D_j;\bm\theta_0)\Big]^{-1}\frac{1}{\sqrt{N_b}}\sum_{j=1}^b\nabla L_{D_j}(\bm\theta_0)+o_p(1).
\end{aligned}
\label{th3proof7}
\end{equation}
By the equation (\ref{th3proof7}) and the central limit theorem, we have
$\sqrt{N_b}\left(\tilde{\boldsymbol{\theta}}_b^p-\boldsymbol{\theta}_0\right) \xrightarrow{\mathrm{D}} \mathcal{N}\left(\bm 0, \boldsymbol{\Sigma}_0\right).$

\end{proof}
\par
\noindent
{\bf Proof of Theorem 3}:
\begin{proof}
Define
\begin{eqnarray}
l_{D^*}(\bm\theta)=
\frac{1}{N_b}\sum_{i=1}^{N_b}V_q(y_i\bar{\bm{x}}_i^T\bm{\theta})+ \frac{\lambda}{2}\bm\theta^T\bm\theta.
\label{th4proof1}
\end{eqnarray}
Recall that
\begin{equation}
\begin{aligned}
Q(\bm\theta)
=&l_{D_b}(\tilde{\boldsymbol{\theta}}_{b-1}^p)+ 
\frac{1}{N_b}(\boldsymbol{\theta}-\tilde{\boldsymbol{\theta}}_{b-1}^p)^T\nabla L_{D_b}(\tilde{\bm\theta}_{b-1}^p)+
\frac{1}{2N_b}(\boldsymbol{\theta}-\tilde{\boldsymbol{\theta}}_{b-1}^p)^T\sum_{j=1}^{b}\bar H_{D_j}(\tilde{\boldsymbol{\theta}}_{j-1}^p)(\boldsymbol{\theta}-\tilde{\boldsymbol{\theta}}_{b-1}^p)\\
&+\frac{\rho}{2 N_b}\|\boldsymbol{\theta}\|^2+\frac{\boldsymbol{\xi}^T \boldsymbol{\theta}}{N_b},
\end{aligned}
\label{th4proof2}
\end{equation}
where
$D^*=\{D_1, D_2, \cdots, D_b\}$.
\\
If we collect all the data $D^*$ without considering privacy protection, an estimator of the parameter $\bm{\theta}$ can obtained by minimizing the loss function (\ref{th4proof1}). That is, $\tilde{\boldsymbol{\theta}}=\arg \min _{\boldsymbol{\theta}}l_{D^*}(\bm\theta)$.
\\
In addition, following the equation (\ref{Q-2}), we can achieve the proposed estimator $\tilde{\boldsymbol{\theta}}_b^p$ by minimizing the loss function (\ref{th4proof2}),  namely, $\tilde{\boldsymbol{\theta}}_b^p=\arg \min _{\boldsymbol{\theta}}Q(\bm\theta)$.
\\
Taking the second-order Taylor series
expansion of $l_{D^*}(\bm\theta)$ around $\tilde{\boldsymbol{\theta}}_{b-1}^p$, we have 
\begin{equation}
\begin{aligned}
l_{D^*}(\bm\theta)
=& l_{D^*}(\tilde{\boldsymbol{\theta}}_{b-1}^p)+ \frac{1}{N_b}\nabla L_{D^*}(\tilde{\bm\theta}_{b-1}^p)(\boldsymbol{\theta}-\tilde{\boldsymbol{\theta}}_{b-1}^p)+
\frac{1}{2N_b}(\boldsymbol{\theta}-\tilde{\boldsymbol{\theta}}_{b-1}^p)^T\bar H_{D^*}(\tilde{\boldsymbol{\theta}}_{b-1}^p)(\boldsymbol{\theta}-\tilde{\boldsymbol{\theta}}_{b-1}^p)\\
&+O_p(\|\boldsymbol{\theta}-\tilde{\boldsymbol{\theta}}_{b-1}^p\|^3),
\end{aligned}
\label{th4proof3}
\end{equation}
where $ \nabla L_{D^*}(\tilde{\bm\theta}_{b-1}^p)=
\sum_{m=1}^M\nabla L_{D^*}^m(\tilde{\bm\theta}_{b-1}^p)$ and 
$ \bar H_{D^*}(\tilde{\bm\theta}_{b-1}^p)=
\sum_{m=1}^M \bar H_{D^*}^m(\tilde{\bm\theta}_{b-1}^p)$.
\\
Further, 
\begin{equation}
\begin{aligned}
l_{D^*}(\bm\theta)
=&l_{D^*}(\tilde{\boldsymbol{\theta}}_{b-1}^p)+ \frac{1}{N_b}\sum_{j=1}^b\nabla L_{D_j}(\tilde{\bm\theta}_{b-1}^p)(\boldsymbol{\theta}-\tilde{\boldsymbol{\theta}}_{b-1}^p)+
\frac{1}{2N_b}(\boldsymbol{\theta}-\tilde{\boldsymbol{\theta}}_{b-1}^p)^T\sum_{j=1}^b\bar H_{D_j}(\tilde{\boldsymbol{\theta}}_{b-1}^p)(\boldsymbol{\theta}-\tilde{\boldsymbol{\theta}}_{b-1}^p)\\
&+O_p(\|\boldsymbol{\theta}-\tilde{\boldsymbol{\theta}}_{b-1}^p\|^3).
\end{aligned}
\label{th4proof4}
\end{equation}
By the Lipschitz continuity, there exists $B_L(D_j)>0$ such that
\begin{equation}\nonumber
\begin{aligned}
\|\nabla L_{D_j}^m(\tilde{\bm\theta}_{b-1}^p)-\nabla L_{D_j}^m(\bm\theta_0)\| \leqslant B_L(D_j)\|\tilde{\boldsymbol{\theta}}_{b-1}^p-\boldsymbol{\theta}_0\|.
\end{aligned}
\end{equation}
It follows that 
\begin{equation}
\begin{aligned}
\nabla L_{D_j}^m(\tilde{\bm\theta}_{b-1}^p)-\nabla L_{D_j}^m(\bm\theta_0)
=O_p(n_j^{\scriptscriptstyle(m)}\|\tilde{\boldsymbol{\theta}}_{b-1}^p-\boldsymbol{\theta}_0\|).
\end{aligned}
\label{th4proof6}
\end{equation}
In addition, it is easy to see that
$\nabla L_{D_j}^m(\bm\theta_0)=o(n_j^{\scriptscriptstyle(m)})$.
Therefore, we obtain
\begin{equation}
\begin{aligned}
\nabla L_{D_j}^m(\tilde{\bm\theta}_{b-1}^p)
=O_p(n_j^{\scriptscriptstyle(m)}\|\tilde{\boldsymbol{\theta}}_{b-1}^p-\boldsymbol{\theta}_0\|)
+ o(n_j^{\scriptscriptstyle(m)}).
\end{aligned}
\label{th4proof7}
\end{equation}
This result of the theorem 3 implies that
\begin{equation}
\begin{aligned}
\|\tilde{\boldsymbol{\theta}}_{b-1}^p-\boldsymbol{\theta}_0\|=O_p\big(1/ \sqrt{N_{b-1}}\big). 
\end{aligned}
\label{th4proof8}
\end{equation}
Note that 
$n_j=\sum_{m=1}^Mn_j^{\scriptscriptstyle(m)}$, $N_k=\sum_{j=1}^kn_j, k=1,\cdots, b$,
$ \nabla L_{D_j}(\tilde{\bm\theta}_{b-1}^p)=
\sum_{m=1}^M\nabla L_{D_j}^m(\tilde{\bm\theta}_{b-1}^p)$.
Combining (\ref{th4proof7}) and (\ref{th4proof8}) yields
\begin{equation}\nonumber
\begin{aligned}
\sum_{j=1}^{b-1}\nabla L_{D_j}(\tilde{\bm\theta}_{b-1}^p)=O_p\big(\sqrt{N_{b-1}}\big) + o(N_{b-1})=O_p\big(\sqrt{N_{b-1}}\big).
\end{aligned}
\end{equation}
This result shows that the gradient of the loss function evaluated at $\tilde{\boldsymbol{\theta}}_{b-1}^p, \sum_{j=1}^{b-1}\nabla L_{D_j}(\tilde{\bm\theta}_{b-1}^p)$, is of order 
$O_p\big(\sqrt{N_{b-1}}\big)$.
\\
Thus, 
\begin{equation}
\begin{aligned}
\frac{1}{N_b}\sum_{j=1}^{b-1}\nabla L_{D_j}(\tilde{\bm\theta}_{b-1}^p)(\boldsymbol{\theta}-\tilde{\boldsymbol{\theta}}_{b-1}^p)=O_p\Big(\frac{\sqrt{N_{b-1}}}{N_b}\|\boldsymbol{\theta}-\tilde{\boldsymbol{\theta}}_{b-1}^p\|\Big).
\end{aligned}
\label{th4proof10}
\end{equation}
Similar to the equation (\ref{th4proof6}), by the Lipschitz continuity, we have
\begin{equation}
\begin{aligned}
\bar H_{D_j}^m(\tilde{\bm\theta}_{b-1}^p)-\bar H_{D_j}^m(\tilde{\bm\theta}_{j-1}^p)
=O_p(n_j^{\scriptscriptstyle(m)}\|\tilde{\boldsymbol{\theta}}_{b-1}^p-\tilde{\boldsymbol{\theta}}_{j-1}^p\|).
\end{aligned}
\label{th4proof11}
\end{equation}
Let $\tilde{\boldsymbol{\theta}}_{0}^p$ denote a consistent estimator of $\boldsymbol{\theta}_1^0$, it follows that
\begin{equation}
\begin{aligned}
\frac{1}{2N_b}\Big[\sum_{j=1}^{b-1}\bar H_{D_j}(\tilde{\boldsymbol{\theta}}_{b-1}^p)
-\sum_{j=1}^{b-1}\bar H_{D_j}(\tilde{\boldsymbol{\theta}}_{j-1}^p)\Big]
=\frac{1}{N_b}O_p\Big(\sum_{j=1}^{b-1}\frac{n_j}{\sqrt{N_{j-1}}}\Big)
\approx O_p\Big(\frac{1}{\sqrt{N_{b}}}\Big)
\end{aligned}
\label{th4proof12}
\end{equation}
Thus, combining(\ref{th4proof2}), (\ref{th4proof4}), (\ref{th4proof10}) and (\ref{th4proof12}), we have
\begin{equation}
\begin{aligned}
Q(\bm\theta)
=&l_{D^*}(\bm\theta)+ O_p\Big(\frac{\sqrt{N_{b-1}}}{N_b}\|\boldsymbol{\theta}-\tilde{\boldsymbol{\theta}}_{b-1}^p\|\Big)+ O_p\Big(\frac{1}{\sqrt{N_{b}}}\|\boldsymbol{\theta}-\tilde{\boldsymbol{\theta}}_{b-1}^p\|^2\Big)\\
&+O_p(\|\boldsymbol{\theta}-\tilde{\boldsymbol{\theta}}_{b-1}^p\|^3) +\frac{\rho}{2 N_b}\|\boldsymbol{\theta}\|^2+\frac{\boldsymbol{\xi}^T \boldsymbol{\theta}}{N_b}.
\end{aligned}
\label{th4proof13}
\end{equation}
By the definition $\tilde{\boldsymbol{\theta}}_{b}^p$, we know
$Q(\tilde{\boldsymbol{\theta}}_{b}^p)\le Q(\bm\theta_0)$. That is,
\begin{equation}\nonumber
\begin{aligned}
&l_{D^*}(\tilde{\boldsymbol{\theta}}_{b}^p)+ O_p\Big(\frac{\sqrt{N_{b-1}}}{N_b}\|\tilde{\boldsymbol{\theta}}_{b}^p-\tilde{\boldsymbol{\theta}}_{b-1}^p\|\Big)+ O_p\Big(\frac{1}{\sqrt{N_{b}}}\|\tilde{\boldsymbol{\theta}}_{b}^p-\tilde{\boldsymbol{\theta}}_{b-1}^p\|^2\Big)\\
&+O_p(\|\tilde{\boldsymbol{\theta}}_{b}^p-\tilde{\boldsymbol{\theta}}_{b-1}^p\|^3)+\frac{\rho}{2 N_b}\|\tilde{\boldsymbol{\theta}}_{b}^p\|^2+\frac{\boldsymbol{\xi}^T\tilde{\boldsymbol{\theta}}_{b}^p}{N_b}\\
&\le
l_{D^*}(\bm\theta_0)+ O_p\Big(\frac{\sqrt{N_{b-1}}}{N_b}\|\bm\theta_0-\tilde{\boldsymbol{\theta}}_{b-1}^p\|\Big)+ O_p\Big(\frac{1}{\sqrt{N_{b}}}\|\bm\theta_0-\tilde{\boldsymbol{\theta}}_{b-1}^p\|^2\Big)\\
&+O_p(\|\bm\theta_0-\tilde{\boldsymbol{\theta}}_{b-1}^p\|^3)+\frac{\rho}{2 N_b}\|\boldsymbol{\theta}_0\|^2+\frac{\boldsymbol{\xi}^T \boldsymbol{\theta}_0}{N_b}
\end{aligned}
\label{th4proof14}
\end{equation}
Recall that 
$l_{D^*}(\bm\theta)=
\frac{1}{N_b}\sum_{i=1}^{N_b}V_q(y_i\bar{\bm{x}}_i^T\bm{\theta})+ \frac{\lambda}{2}\bm\theta^T\bm\theta$. Therefore,
\begin{equation}
\begin{aligned}
0\le\varepsilon_{E}&=E_{\bm{X} Y}[V_q\{Y f_{\tilde{\bm\theta}_b^p}(\bm X)\}]-E_{\bm X Y}[V_q\{Y f_{\bm\theta_0}(\bm X)\}]\\
&\le B_1+B_2+B_3+B_4+B_5+B_6.
\end{aligned}
\label{th4proof15}
\end{equation}
where
\begin{equation}\nonumber
\begin{aligned}
&B_1= \frac{\lambda}{2}(\bm\|\bm\theta_0\|^2-\|\tilde{\bm\theta}_b^p\|^2).
\\
&B_2= O_p\Big(\frac{\sqrt{N_{b-1}}}{N_b}\|\bm\theta_0-\tilde{\boldsymbol{\theta}}_{b-1}^p\|\Big)- O_p\Big(\frac{\sqrt{N_{b-1}}}{N_b}\|\tilde{\boldsymbol{\theta}}_{b}^p-\tilde{\boldsymbol{\theta}}_{b-1}^p\|\Big).
\\
&B_3= O_p\Big(\frac{1}{\sqrt{N_{b}}}\|\bm\theta_0-\tilde{\boldsymbol{\theta}}_{b-1}^p\|^2\Big)- O_p\Big(\frac{1}{\sqrt{N_{b}}}\|\tilde{\boldsymbol{\theta}}_{b}^p-\tilde{\boldsymbol{\theta}}_{b-1}^p\|^2\Big).
\\
&B_4= O_p(\|\bm\theta_0-\tilde{\boldsymbol{\theta}}_{b-1}^p\|^3)-O_p(\|\tilde{\boldsymbol{\theta}}_{b}^p-\tilde{\boldsymbol{\theta}}_{b-1}^p\|^3).
\\
&B_5=\frac{\rho}{2N_b}(\bm\|\bm\theta_0\|^2-\|\tilde{\bm\theta}_b^p\|^2).
\\
&B_6=\frac{\bm\xi^T}{N_b}(\bm\theta_0-\tilde{\bm\theta}_b^p).
\end{aligned}
\label{th4proof16}
\end{equation}
Note that $\bm\theta_0=O(1)$, $\lambda=O(1)$, $\rho=O(1)$,
$
\|\tilde{\boldsymbol{\theta}}_{b-1}^p-\boldsymbol{\theta}_0\|=O_p\big(1/ \sqrt{N_{b-1}}\big)
$
and
$
\|\tilde{\boldsymbol{\theta}}_{b}^p-\boldsymbol{\theta}_0\|=O_p\big(1/ \sqrt{N_{b}}\big).
$
Furthermore,
\begin{equation}
\begin{aligned}
B_1= \frac{\lambda}{2}(\bm\|\bm\theta_0\|^2-\|\tilde{\bm\theta}_b^p\|^2)
\le 
\frac{\lambda}{2}(\|\bm\theta_0-\tilde{\bm\theta}_b^p\|)(\bm\|\bm\theta_0\|+\|\tilde{\bm\theta}_b^p\|)
=\lambda O_p\big(\frac{\|\bm\theta_0\|+O_p\big(1/ \sqrt{N_{b}}\big)}{\sqrt{N_{b}}}\big)
= O_p\big(\frac{1}{\sqrt{N_{b}}}\big).
\end{aligned}
\label{th4proofB1}
\end{equation}
\begin{equation}
\begin{aligned}
B_2= O_p\Big(\frac{\sqrt{N_{b-1}}}{N_b}\|\bm\theta_0-\tilde{\boldsymbol{\theta}}_{b-1}^p\|\Big)- O_p\Big(\frac{\sqrt{N_{b-1}}}{N_b}\|\tilde{\boldsymbol{\theta}}_{b}^p-\tilde{\boldsymbol{\theta}}_{b-1}^p\|\Big)
\le
O_p\Big(\frac{1}{\sqrt{N_b}}\|\bm\theta_0-\tilde{\boldsymbol{\theta}}_{b}^p\|\Big)
= O_p\big(\frac{1}{N_{b}}\big).
\end{aligned}
\label{th4proofB2}
\end{equation}
\begin{equation}
\begin{aligned}
B_3&= O_p\Big(\frac{1}{\sqrt{N_{b}}}\|\bm\theta_0-\tilde{\boldsymbol{\theta}}_{b-1}^p\|^2\Big)- O_p\Big(\frac{1}{\sqrt{N_{b}}}\|\tilde{\boldsymbol{\theta}}_{b}^p-\tilde{\boldsymbol{\theta}}_{b-1}^p\|^2\Big)\\
&\le
O_p\Big(\frac{1}{\sqrt{N_{b}}}\|\bm\theta_0-\tilde{\boldsymbol{\theta}}_{b}^p\|\big(\|\bm\theta_0-\tilde{\boldsymbol{\theta}}_{b-1}^p\|+\|\tilde{\boldsymbol{\theta}}_{b}^p-\tilde{\boldsymbol{\theta}}_{b-1}^p\|\big)\Big)\\
&= O_p\big(\frac{1}{N_{b}\sqrt{N_{b-1}}}\big).
\end{aligned}
\label{th4proofB3}
\end{equation}
\begin{equation}
\begin{aligned}
B_4&= O_p\big(\|\bm\theta_0-\tilde{\boldsymbol{\theta}}_{b-1}^p\|^3)-O_p(\|\tilde{\boldsymbol{\theta}}_{b}^p-\tilde{\boldsymbol{\theta}}_{b-1}^p\|^3\big)\\
&\le
O_p\big(\|\bm\theta_0-\tilde{\boldsymbol{\theta}}_{b}^p\|(\|\bm\theta_0-\tilde{\boldsymbol{\theta}}_{b-1}^p\|^2+\|\bm\theta_0-\tilde{\boldsymbol{\theta}}_{b-1}^p\|\|\tilde{\boldsymbol{\theta}}_{b}^p-\tilde{\boldsymbol{\theta}}_{b-1}^p\|+\|\tilde{\boldsymbol{\theta}}_{b}^p-\tilde{\boldsymbol{\theta}}_{b-1}^p\|^2)\big)\\
&= O_p\big(\frac{1}{N_{b-1}\sqrt{N_{b}}}\big).
\end{aligned}
\label{th4proofB4}
\end{equation}
Similar to $B_1$, 
\begin{equation}
\begin{aligned}
B_5=\frac{\rho}{2N_b}(\bm\|\bm\theta_0\|^2-\|\tilde{\bm\theta}_b^p\|^2)
\le 
\frac{\rho}{2N_b}(\|\bm\theta_0-\tilde{\bm\theta}_b^p\|)(\bm\|\bm\theta_0\|+\|\tilde{\bm\theta}_b^p\|)
= O_p\big(\frac{1}{N_b\sqrt{N_{b}}}\big).
\end{aligned}
\label{th4proofB5}
\end{equation}
Since $\tilde{\bm\theta}_b^p$ is a consistent estimator of $\bm\theta_0$,
\begin{equation}
\begin{aligned}
B_6=\frac{\bm\xi^T}{N_b}(\bm\theta_0-\tilde{\bm\theta}_b^p)=o_p(1).
\end{aligned}
\label{th4proofB6}
\end{equation}
Combining (\ref{th4proof15}), (\ref{th4proofB1}), (\ref{th4proofB2}), (\ref{th4proofB3}), (\ref{th4proofB4}), (\ref{th4proofB5}) and (\ref{th4proofB6}), we have 
$\varepsilon_{E}\xrightarrow{\mathrm{P}} 0,$ as $N_b\rightarrow\infty.$ Thus, by the lemma 2, the proof is completed.

\end{proof}

\end{document}